\documentclass{article}

\usepackage{arxiv}

\usepackage[utf8]{inputenc} 
\usepackage[T1]{fontenc}    
\usepackage{hyperref}       
\usepackage{url}            
\usepackage{booktabs}       
\usepackage{amsfonts}       
\usepackage{nicefrac}       
\usepackage{microtype}      
\usepackage{lipsum}
\usepackage{graphicx}
\graphicspath{ {./images/} }

\usepackage[table]{xcolor} 
\usepackage{multirow}
\def\thickhline{\noalign{\hrule height1.3pt}} 
\def\narrowhline{\noalign{\hrule height.4pt}}
\usepackage{subcaption} 

\title{Leveraging Large Language Models for enzymatic reaction prediction and characterization}

\author{
 Lorenzo Di Fruscia \\
  Department of Intelligent Systems\\
  Delft University of Technology\\
  Delft 2629 HZ, The Netherlands \\
  \texttt{l.difruscia@tudelft.nl} \\
   \And
 Jana M. Weber \\
  Department of Intelligent Systems\\
  Delft University of Technology\\
  Delft 2629 HZ, The Netherlands \\
  \texttt{j.m.weber@tudelft.nl} \\
}

\begin{document}
\maketitle
\begin{abstract}
Predicting enzymatic reactions is crucial for applications in biocatalysis, metabolic engineering, and drug discovery, yet it remains a complex and resource-intensive task. Large Language Models (LLMs) have recently demonstrated remarkable success in various scientific domains, e.g., through their ability to generalize knowledge, reason over complex structures, and leverage in-context learning strategies. In this study, we systematically evaluate the capability of LLMs, particularly the Llama-3.1 family (8B and 70B), across three core biochemical tasks: Enzyme Commission number prediction, forward synthesis, and retrosynthesis. We compare single-task and multitask learning strategies, employing parameter-efficient fine-tuning via LoRA adapters. Additionally, we assess performance across different data regimes to explore their adaptability in low-data settings.
Our results demonstrate that fine-tuned LLMs capture biochemical knowledge, with multitask learning enhancing forward- and retrosynthesis 
predictions by leveraging shared enzymatic information. We also identify key limitations, for example challenges in hierarchical EC classification schemes, highlighting areas for further improvement in LLM-driven biochemical modeling.

\end{abstract}

\section{Introduction}

Biochemistry plays a fundamental role in nearly every aspect of daily life, from  medicine development to food production, from the creation of fuels to personal care items, significantly contributing to improved quality of life. Developing  novel biocatalysts and discovering and optimizing biochemical reactions hold immense promise for addressing global challenges. However, these discoveries are inherently complex, requiring a deep understanding of enzyme-substrate relationships, and they remain experimentally expensive and time-intensive \cite{wiltschiEnzymesRevolutionizeBioproduction2020, alcantaraBiocatalysisKeySustainable2022, sheldonGreenChemistryBiocatalysis2024}.

Machine learning (ML) has transformed many fields, and ML models are increasingly applied to tackle challenges in the molecular sciences. Transformers in particular \cite{vaswaniAttentionAllYou2023}, architectures suited for applications like language translation, sentiment analysis and text completion, have proven to be effective for tasks such as chemical reaction product prediction and molecule optimization \cite{schwallerMolecularTransformerModel2019,pesciullesiTransferLearningEnables2020, irwinChemformerPretrainedTransformer2022}. In biochemistry, several ML models have been tailored for prediction tasks, including approaches where enzymes are represented using natural language \cite{kreutterPredictingEnzymaticReactions2021}, numerical classification schemes \cite{probstBiocatalysedSynthesisPlanning2022, qianGeneralModelPredicting2024}, or amino acid sequences \cite{nanateukamLanguageModelsCan2024}. While these specialized models deliver impressive results, they are typically constrained to specific tasks and require extensive domain-specific data and expertise for their development and for the incorporation of biochemical knowledge.

Recent years have witnessed the emergence of foundation models like Large Language Models (LLMs) \cite{radfordImprovingLanguageUnderstanding2018, radfordLanguageModelsAre2019}, that have found their application in chemistry as well \cite{whiteAssessmentChemistryKnowledge2023}. These transformer-based architectures consist of up to hundreds of billions of parameters and are trained on text corpora comprising trillions of tokens. Despite being trained for next token prediction, these models have shown emergent abilities that were not foreseeable for smaller sized models \cite{weiEmergentAbilitiesLarge2022}: they are capable of more than just completing phrases in natural language, as to some extent they are able to answer questions, understand examples and reason over problems. Foundation models can be capable of solving multiple tasks at once. Building on top of existing LLMs is straightforward to implement and they require relatively little expertise to use, circumventing the need to train a multitude of specialized models. The scientific community is building upon recent discoveries that scaling up LLMs in size and training data leads to promising zero- and few-shot capabilities for \textit{in-context learning} \cite{brownLanguageModelsAre2020}. 

One key problem of learning from context is the high variance in the outputs returned by the model: slight changes in prompts can greatly affect the model performance, ranging from barely above chance, to near state-of-the-art (SOTA) level \cite{zhaoCalibrateUseImproving2021}. Additionally, LLMs may produce made-up or irrelevant content, a phenomenon known as \textit{hallucinations}. To address these instabilities, research has explored advanced prompting strategies such as \textit{Chain-of-Thought} (CoT), a technique that guides the model to break down answers as a series of connected thoughts. By explicitly decomposing complex problems into step-by-step reasoning, CoT reduces output variability and enhances accuracy, particularly for tasks requiring logical progression or multi-step calculations. By acting in a way that mimics human reasoning, CoT showed to improve the reliability of responses and to therewith make LLMs more robust \cite{weiChainofThoughtPromptingElicits2023}.

Another key task adaptation strategy is \textit{fine-tuning}, which modifies the weights of the pretrained model. It offers the advantage of not being constrained by a limited context window for input data, but it typically results in a model specialized for a single task. However, prior research \cite{mosbachFewshotFinetuningVs2023} showed that fine-tuning outperforms in-context learning strategies in both in-domain and out-of-distribution tasks for models of comparable size, with performance gains increasing as more training data becomes available. Fine-tuning limitations in principle include the need for significant training expertise and computational resources, with a reduced reusability compared to in-context learning strategies. These shortcomings are partially mitigated by Parameter-Efficient Fine-Tuning (PEFT) \cite{hanParameterEfficientFineTuningLarge2024, xuParameterEfficientFineTuningMethods2023}. PEFT techniques selectively adjust only a small portion of parameters, leaving the rest unchanged. This approach preserves the base model general-purpose capabilities while adding task-specific expertise in a modular way, enabling greater adaptability to new tasks.

Efforts to integrate LLMs into chemistry generally fall into two distinct categories. The first focuses on building chemistry agents that leverage the LLMs planning abilities to work with task-specific tools and improve reasoning \cite{openaiGPT4TechnicalReport2024}. For instance, in Bran et al. \cite{branChemCrowAugmentingLargelanguage2023}, researchers augmented LLMs by providing access to expert-designed tools for drug discovery, materials design and organic synthesis. 
The second category involves using LLMs directly for downstream tasks such as property prediction, reagent selection and molecule captioning \cite{guoWhatCanLarge2023, jablonkaLeveragingLargeLanguage2024, maikjablonka14ExamplesHow2023, zhengShapingWaterHarvestingBehavior2023}. In Guo et al. \cite{guoWhatCanLarge2023}, they benchmarked LLMs in zero- and few-shot settings, demonstrating their capabilities in explaining, understanding and reasoning over chemistry.
In Jablonka et al. \cite{jablonkaLeveragingLargeLanguage2024}, they show how by fine-tuning GPT family models from OpenAI \cite{openAI}, they easily adapt them to solve various tasks involving classification, regression, inverse design of chemicals, and many more. Their model proved to be useful especially in the low-data regime, where the LLM performed at least as good as the conventional ML models.
Additionally, comprehensive instruction datasets for the chemical and biochemical domains have been introduced \cite{fangMolInstructionsLargeScaleBiomolecular2024, yuLlaSMolAdvancingLarge2024}. These datasets, encompassing millions of examples across applications like molecule generation, name conversion and reaction prediction, enable small fine-tuned LLMs to surpass prompted SOTA LLMs, demonstrating the role of high quality datasets in enhancing performance in molecular domains.

While previous studies mainly focused on the investigation of LLMs for chemical and materials tasks, we are interested in understanding LLMs potential for biochemical reaction characterization, discovery, and optimization. 
We focus on enzymatic reactions represented using SMILES (Simplified Molecular Input Line Entry System) notation \cite{weiningerSMILESChemicalLanguage1988a} for chemicals and EC numbers for enzyme classification. Specifically, we design tasks that test the model’s ability to predict EC numbers, reaction products (forward synthesis), and substrates (retrosynthesis). By introducing a multitask learning setup, we investigate whether training on multiple tasks simultaneously makes use of shared biochemical knowledge compared to single-task fine-tuning. 
Finally, we perform ablation studies to examine the impact of several data regimes and fine-tuning setups on different models' performance.

\section{Methods}
\subsection{Tasks and dataset description}
\label{sec:task_and_dataset}

\paragraph{Task selection}\mbox{}\\
We assemble a representative set of biochemical prediction tasks. The selected tasks are designed to evaluate the capabilities of Large Language Models (LLMs) in understanding and predicting enzymatic reactions, when the chemicals are presented in string format and the enzyme in the EC numerical classification scheme. Specifically:
\begin{itemize}
    \item EC Number Prediction: we assess whether LLMs can accurately assign EC numbers given the substrates and the products of each reaction.
    \item Product Prediction: here we explore the model's ability to predict reaction products given substrates and the EC number associated to the reaction (forward synthesis).
    \item Substrate Prediction: we test the model’s capabilities of predicting substrates based on reaction products and the EC number (retrosynthesis).
\end{itemize}

Given the inherent similarities among the three tasks, we investigate whether the model can improve its performance when trained on all tasks simultaneously, by leveraging shared information in a synergistic manner. To test this, we introduce a multitask (MT) setup, in which a single model is trained concurrently on all three tasks, inspired by what has been done in Yu et al. \cite{yuLlaSMolAdvancingLarge2024}. This setup allows us to evaluate whether a multitask-trained model can outperform individually fine-tuned models for each task (single-task, ST) producing a more general model eventually capable of handling diverse biochemistry tasks involving enzymes. The following sections explain data selection and the data split suitable for both ST and MT experiments. To ensure that the selected tasks are supported by high-quality data, we preprocess the data to minimize biases and data leakage.

\begin{figure}[h!]
    \centering
    \includegraphics[width=0.6\linewidth]{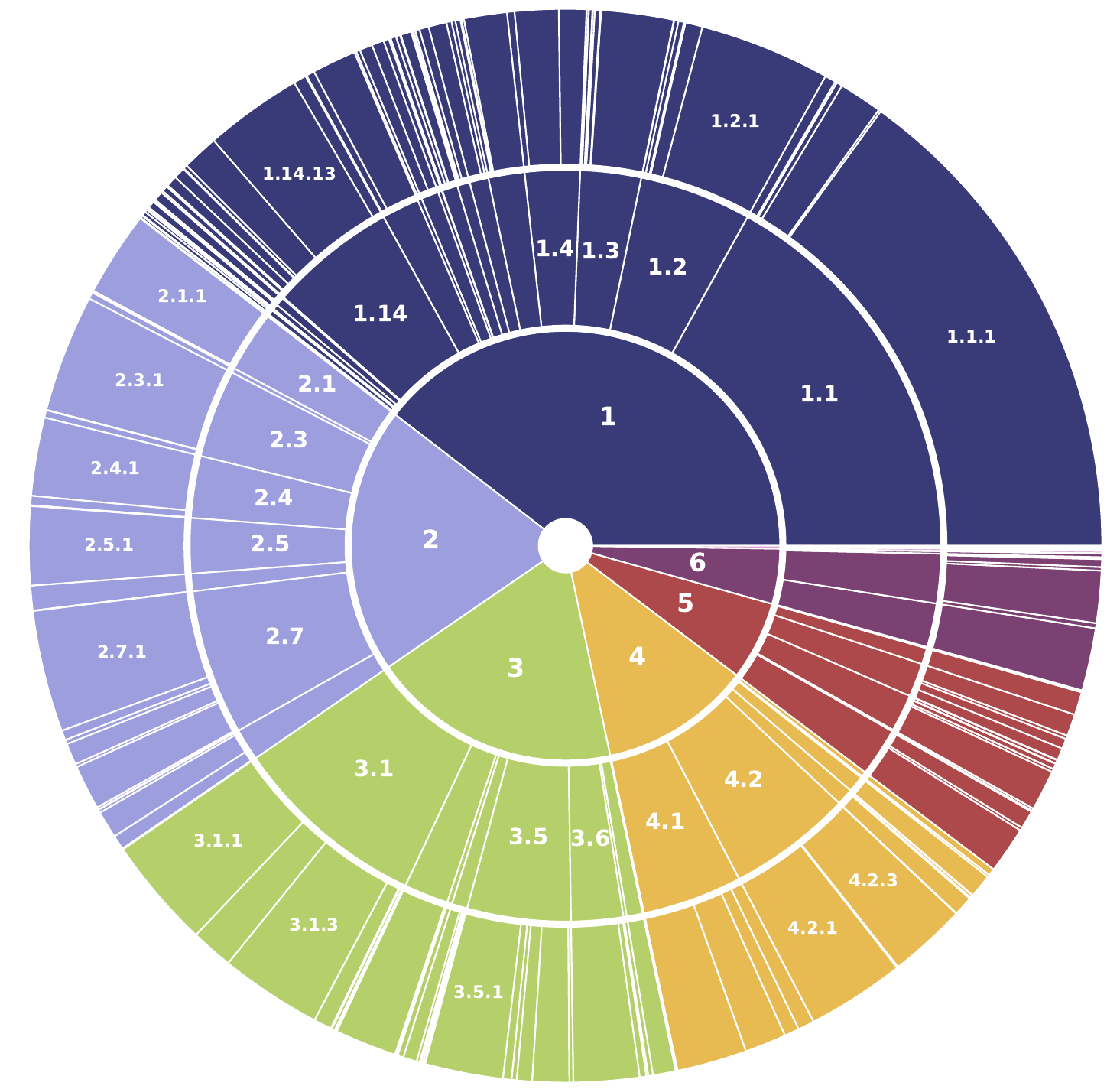}
    \caption{Distributions of samples across EC levels for the BRENDA dataset. The innermost layer represents the main class (EC1 digit), and the middle and outer layers represent levels EC2 and EC3 respectively. The label for enzyme class 7 (\textit{translocases}) is not visible due to the limited data available (<20 samples).}
    \label{fig:brenda_pie}
\end{figure}

\paragraph{Dataset preparation}\mbox{}\\
We make use of the ECREACT dataset curated by Probst et al \cite{probstBiocatalysedSynthesisPlanning2022}. This dataset results from the combination of data coming from four different databases: MetaNetX, Rhea, PathBank and BRENDA \cite{Metanetx,Rhea,PathBank,Brenda}. The authors screened the enzymatic reactions, and determined the corresponding Enzyme Commission (EC) number for each of them. Further processing simplified and generalized the dataset. They removed products also occurring as reactants in the same reaction, co-enzymes, common by-products, and reactions without reactants or multiple or missing products. In each reaction, substrates and products are represented in SMILES, whereas EC numbers are tags for the reaction in the form of a 4-digit tag 'X.X.X.X'. The digits follow a hierarchy, with the first digit (EC1) representing the main class of the enzymatic reaction. From these databases, we only focus on BRENDA, 
for a total of $\textbf{n=8496}$ enzyme-catalyzed reactions covering all seven different EC classes. This is mainly due to computational constraints, as fine-tuning large-scale LLMs on the full dataset without parallelized infrastructure would require several weeks.
The distribution of reactions according to their respective EC numbers is shown in Figure \ref{fig:brenda_pie}. We include all four EC digits (thus up to EC4) in the dataset, but our subsequent analyses will focus on up to sublevel EC3, as many subcategories for EC4 consist of only a single enzyme-substrate example. Class 7 will not be included as well due to the limited sample size for the class ($<20$ samples).

\paragraph{Data splitting}\mbox{}\\
We implement several preprocessing steps: canonicalization of SMILES representations to remove redundant entries, grouping reactions that share the same \textit{{product, EC}} or \textit{{substrate, EC}} pair, but differ in the remaining molecule, and avoidance of task-specific leakage, ensuring that if \textit{e.g.} a reaction appears in forward synthesis, it must not appear in retrosynthesis as well. More details about these steps are reported in Appendix section \ref{Appendix:preprocessing}

\begin{figure}[h!]
    \centering
    \includegraphics[width=0.7\linewidth]{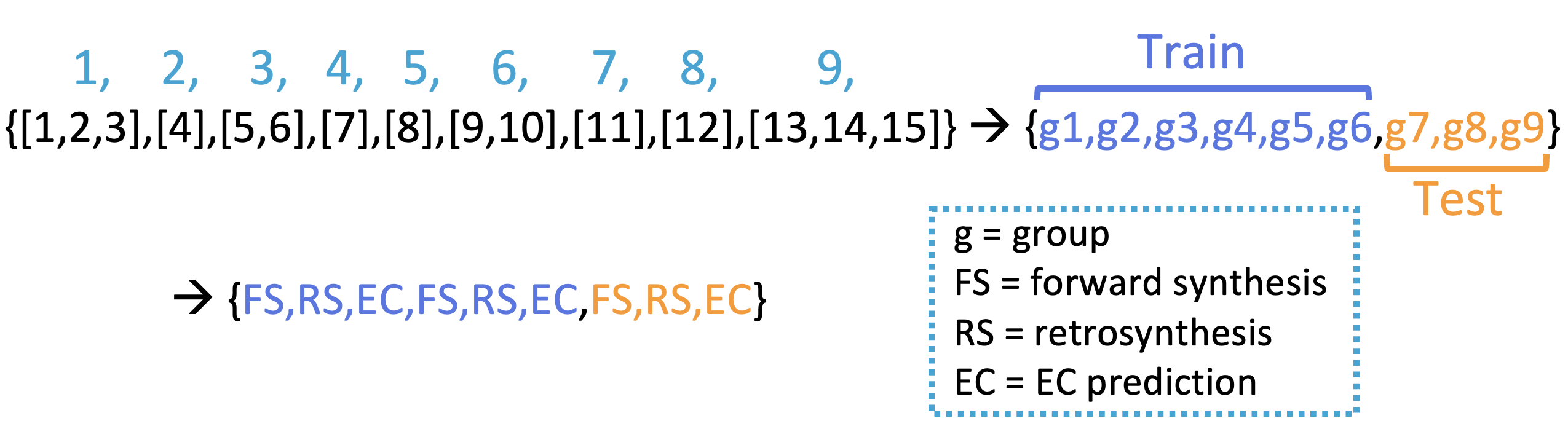}
    \caption{Individual reactions sharing the same $\{product, EC\}$ or $\{substrate, EC\}$ pair are grouped together (here groups are numbered from 1 to 9, first row). The dataset is split into training and test set, while keeping each group intact. Within training and test, each group is assigned to one of the three tasks on a rotating basis to balance the splits. Groups are randomly shuffled at the beginning of the procedure, here we keep indices in order for visual clarity.}
    \label{fig:split}
\end{figure}

These points imply that to maintain dataset integrity, each above-mentioned reaction group is assigned exclusively to one task and one dataset split (either training or test). By addressing these issues preemptively, we also ensure a consistent random dataset split for both single-task and multitask setups, enabling fair comparisons between the two methodologies.
Figure \ref{fig:split} better illustrates this approach.

\begin{figure}[h!]
    \centering
    \includegraphics[width=0.65\linewidth]{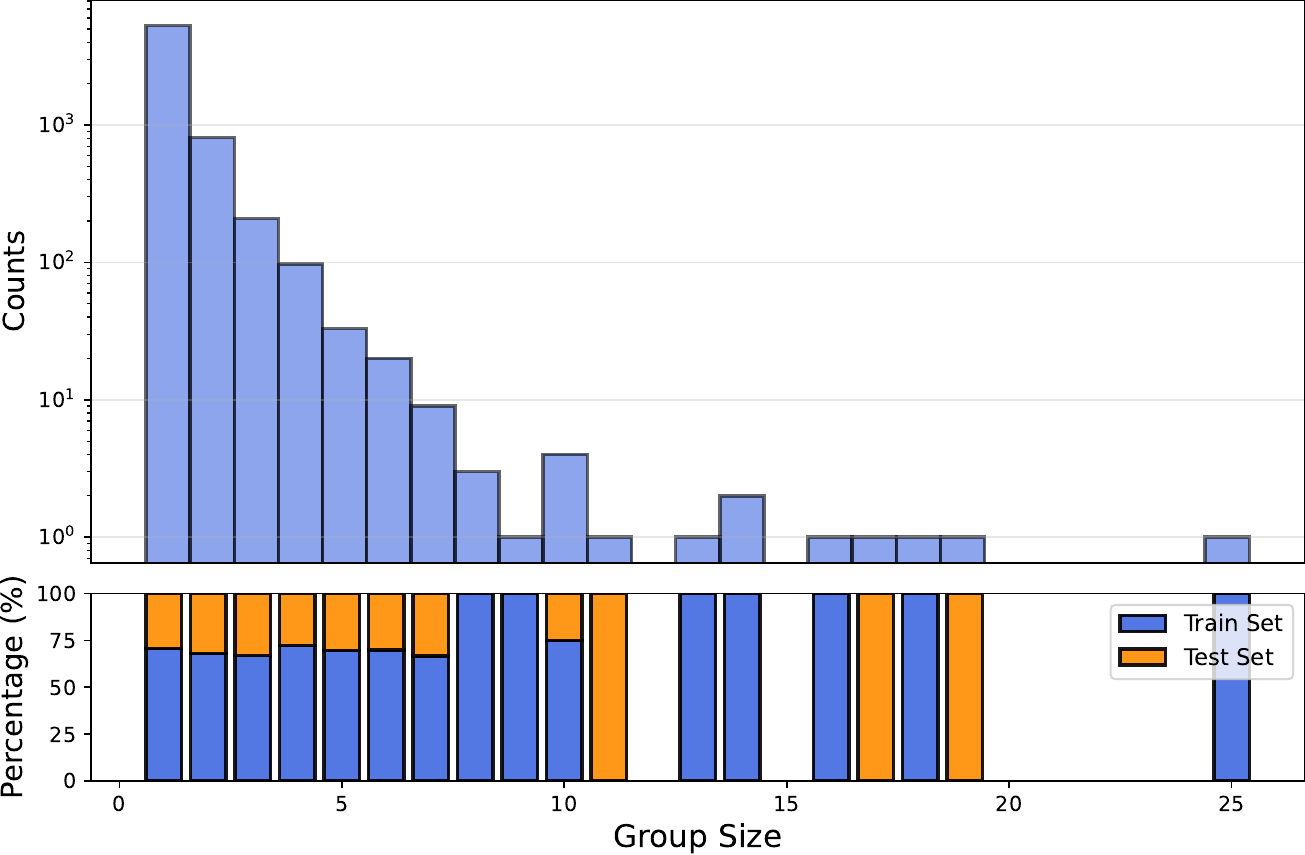}
    \caption{Distribution of reaction groups with repeating substrates and/or products. Unique reactions are included as elements with group size equal to 1. Group sizes with a number of counts $>10$ closely follow the required 70-30 split ratio between train and test set.}
    \label{fig:groups_split}
\end{figure}

We perform a 70-30 train-test split, ensuring that the fraction of groups assigned to each of the two sets maintains a balanced ratio. Of the train set, 10\% is used for validation. 
The above-mentioned preprocessing steps prevent information leakage that could artificially inflate performance metrics, although being public it is likely that the LLM may have had access to this data during its extensive pretraining. Figure \ref{fig:groups_split} illustrates the final distribution of groups across the train and test sets. 

\subsection{LLM interaction and adaptation}
\paragraph{In-Context Learning}\mbox{}\\
In In-Context Learning (ICL), we interact with a LLM solely through prompting. Prompting means giving a set of instructions to the model in natural language in order to make it perform a task: answering, reasoning, story-writing, conversation, tool-access and so on. LLMs are powerful zero-shot learners and can easily adapt to examples to improve their understanding, which is called few-shot prompting \cite{brownLanguageModelsAre2020}.\\
 When interacting with the LLM, each data point is formatted as a conversation between a user and an assistant, as follows:
\begin{itemize}
    \item A general system prompt assigns the model the role of a biochemically knowledgeable assistant.
    \item The user prompt specifies the task, phrasing it in a flexible way to ensure a certain degree of variability. Diverse templates are used in order to prevent overfitting on specific question structures.
    \item The assistant provides an answer, formatted with tagging elements such as \textit{<EC>} or \textit{</EC>}, 
    to enhance consistency and ease of parsing.    
\end{itemize}

\begin{figure}[h!]
    \centering
    \includegraphics[width=0.7\linewidth]{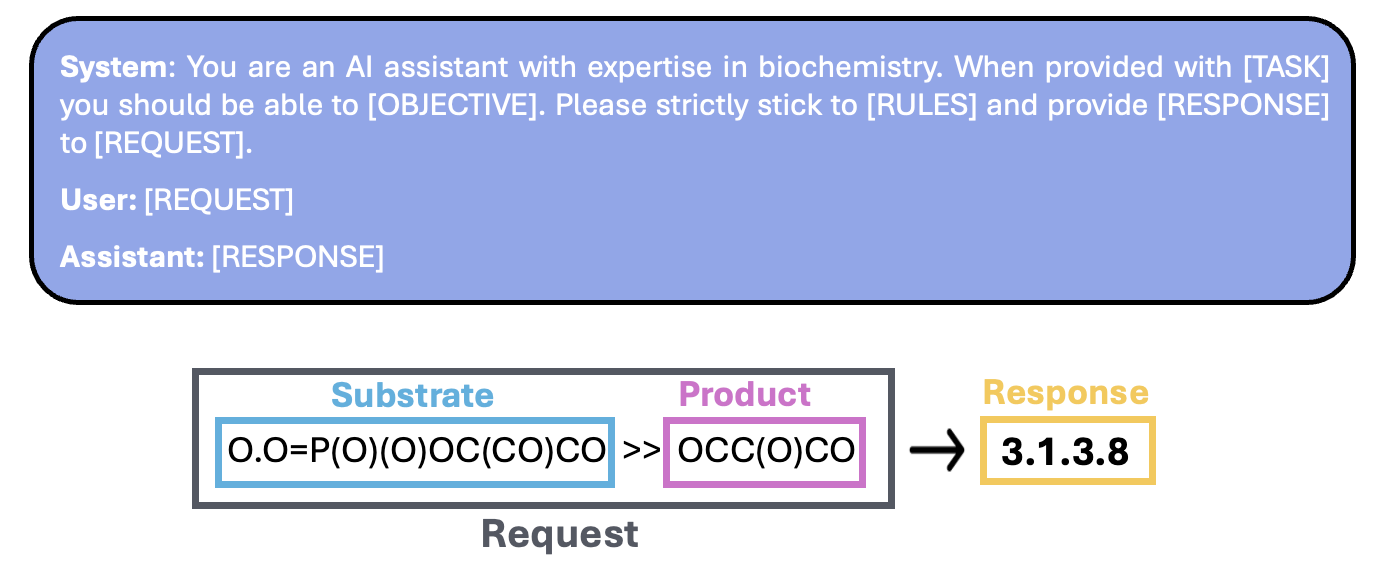}
    \caption{Example of a zero-shot prompt for the EC number prediction task. The model first receives a general prompt with instructions that inform it about the task to perform. The [TASK] here is EC number prediction, and the [OBJECTIVE] is to assign the 4 digits of the EC number given only reactants and product in SMILES notation. After that, the model receives the reaction SMILES from the user as a [REQUEST], and the model associates an EC number to it as the [RESPONSE].}
    \label{fig:prompting}
\end{figure}

A visual example of this is shown in Figure \ref{fig:prompting}.

\paragraph{Fine-tuning}\mbox{}\\
Fine-tuning refers to the process of adapting a pretrained model to perform specific tasks by updating its parameters on a new dataset. This approach allows the model to specialize in a narrower domain while retaining its general pretrained knowledge. To fine-tune our models efficiently, we use Parameter-Efficient Fine-Tuning (PEFT) techniques, specifically Low-Rank Adaptation (LoRA) \cite{huLoRALowRankAdaptation2021}. LoRA allows fine-tuning by updating only a small subset of the model's parameters, significantly reducing computational demands. Instead of directly updating a weight matrix of the pretrained model $W\in \mathbb{R}^{n\times m}$, LoRA models the update as $\Delta W = A\cdot B$, where $ A \in \mathbb{R}^{n \times r} $ and $ B \in \mathbb{R}^{r \times m} $ are matrices with a rank $ r \ll \min(n, m)$. The rank determines the size of the two matrices, and during forward passes the effective weight matrix becomes

\begin{equation}
    W' = W + \Delta W = W + A \cdot B
\end{equation} 

\begin{figure}[h!]
    \centering
    \includegraphics[width=0.31\linewidth]{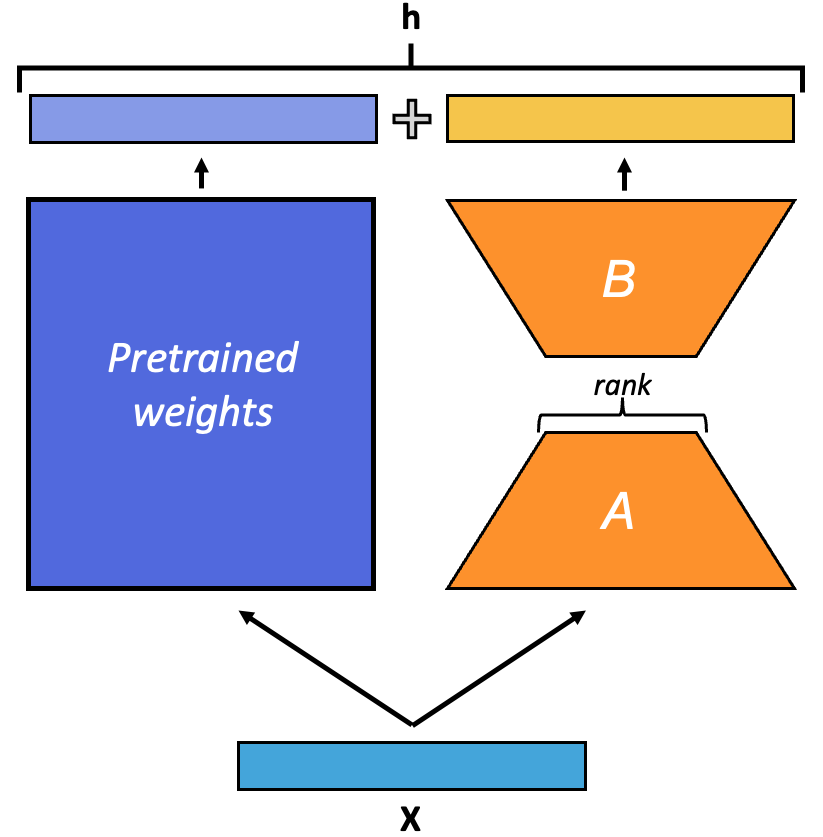}
    \caption{Illustration of LoRA framework. The input vector \textit{x} is passed through both the frozen weight matrix of the pretrained model, and the LoRA head. After both blocks process the input, the two representations are summed together to obtain a new representation $h$.}
    \label{fig:lora}
\end{figure}

The small rank is what ensures that $A$ and $B$ contain far fewer parameters than $W$ and this drastically reduces memory footprint and fine-tuning time. An illustration of the algorithm is shown in Figure \ref{fig:lora}.
While the model can be loaded in a quantized format for efficient memory usage, fine-tuning occurs on a limited percentage of parameters that are stored in full/half precision. This approach has shown to yield performance levels close to those of full model fine-tuning, while maintaining the model's general reasoning abilities and core capabilities \cite{huLoRALowRankAdaptation2021, dettmersQLoRAEfficientFinetuning2023}.

The use of LoRA adapters is particularly advantageous for LLMs like ours. These adapters can be “plugged in” for domain-specific tasks and subsequently removed to revert to the base model, which remains unaffected by fine-tuning, thereby keeping computational costs under control.

\paragraph{Model selection}\mbox{}\\
In selecting a model for our biochemical prediction tasks, our primary selection criteria are:
\begin{itemize}
    \item Power: the model's ability to handle complex tasks and achieve high accuracy;
    \item Flexibility: its ability to tackle diverse tasks both in in-context learning and fine-tuning settings;
    \item Efficiency: the model's computational cost-effectiveness, particularly in resource-constrained environments.
\end{itemize}

We aimed to use a LLM that balances computational power with flexibility, ensuring it can be customized for specialized biochemical applications. We prioritize general-purpose LLMs to evaluate their adaptability and scalability across multiple biochemical tasks.
Equally important was choosing an open-source model to facilitate accessibility and enable further development by other researchers. Given these requirements, we selected models from Meta AI’s Llama 3.1 family \cite{grattafioriLlama3Herd2024}, specifically the 8B and 70B parameter versions. The smaller 8B model offers a trade-off between efficiency and flexibility for exploratory or lower-resource settings, while the 70B model provides greater power. Further, we employed the \textit{instruct} versions of these models, both for in-context learning and fine-tuning. These variants are fine-tuned on instruction-response pairs, helping them generate responses that align with the given instructions. Lastly, we utilize both base models in the 4-bit quantized format to reduce computational costs and inference time.

\subsection{Evaluation metrics}
For the EC prediction task, a prediction is correct if the digits match exactly those of the ground truth. If only the first digit is correct, the model correctly predicted the EC class. If the first two digits are correct, the prediction is correct up to digit EC2, and so on. For the accuracy, we always compute the macro-average to show performance across classes, treating each class as equally important. Additionally, for the main class we report F1 score, precision and recall, as they help provide a more complete picture especially with imbalanced datasets such as ours.
To evaluate product and substrate predictions, we categorize predicted SMILES strings into five distinct groups:

\begin{itemize}
\item Canonical Match (CM): the predicted SMILES string exactly matches the target;
\item Non-Canonical Match (NCM): the prediction matches the target structure but in a non-canonical form;
\item Canonical Valid (CV): the prediction is wrong, but represents a chemically valid molecule in canonical form;
\item Non-Canonical Valid (NCV): The prediction is wrong but chemically valid in a non-canonical form;
\item Invalid (I): the prediction is not a valid SMILES string, either because chemically implausible or incorrectly formatted.
\end{itemize}

For \textit{Valid} chemicals, we additionally examine molecular similarity to determine the potential relevance of the generated SMILES, using the Tanimoto similarity coefficient after computing Daylight fingerprints \cite{DaylightFingerprint} for each molecule. A Tanimoto similarity $>0.85$ is often considered indicative of structurally similar molecules, suggesting that even incorrect predictions may still be chemically meaningful. High similarity scores could for example suggest that the LLM-generated molecule might serve as an alternative substrate in retrosynthetic applications, potentially offering novel biochemical insights. 
It is important to note that the SOTA results that we mention are taken from existing studies and are based on models trained on the entire ECREACT dataset, which comprises unique $n=62222$ enzymatic reactions aggregated from four different databases. In contrast, our experiments are conducted using only the reactions from the BRENDA database.  While this difference in training data size limits direct comparisons with SOTA models, our setup allows for a holistic experimental design within reasonable computational limits. While this limits the comparison to certain extent, it allows us to focus on a single well-curated database, we can systematically evaluate different model sizes, fine-tuning strategies, and data regimes, while still capturing a diverse range of enzymatic reactions. All results are averaged over $N=3$ experiments to provide robust performance metrics, with standard deviations reported where applicable.

\subsection{Fine-tuning setup}

All models are trained with a learning rate $lr=0.002$ using a linear decay scheduler, and $\{\alpha=32, r=16\}$ for the LoRA adapter. We explore two new LoRA setups in addition to the default one, to evaluate the trade-off between fine-tuning parameter count and model performance. Here we list them, including in parenthesis the number of trainable parameters and their percentage with respect to the pretrained, base model:
\begin{itemize}
    \item LoRA \textbf{light} (6.8M, 0.09$\%$ for the 8B, 32.8M, 0.05$\%$ for the 70B): we only fine-tune the query and key matrices within the attention modules $[q_{proj}, k_{proj}]$.
    \item LoRA \textbf{attention} (13.6M, 0.17$\%$ for the 8B, 65.5M, 0.09$\%$ for the 70B): we extend fine-tuning to all matrices within the attention mechanism $[q_{proj}, k_{proj}, v_{proj}, o_{proj}]$. 
    \item LoRA (41.9M, 0.52$\%$ for the 8B, 207M, 0.29$\%$ for the 70B): this is the basic setting and the one used throughout the paper. The adapter tunes all the attention modules and the feed-forward networks (FFN).
\end{itemize}

\section{Results and discussion}
In this section, we present the results of fine-tuning the selected Llama models. The analysis encompasses ST and MT setups, along with experiments designed to evaluate performance in low-data regimes and across different fine-tuning schemes. For each task, the performance is compared against baselines.

\subsection{Single-task fine-tuning}

Llama-3.1 models accurately predict the the highest level of EC number classification, yet show a decline when tasked with the second and third digit. In a single-task settup, the Llama-3.1 model family exhibits some difficulties with exact product and substrate prediction tasks. Interestingly, we find that reasonably large percentages of uncorrect predictions show a high Tanimoto similarity with the correct predictions, which can potentially still be useful in biochemical workflows.

\begin{figure}[h!]
    \centering
    \includegraphics[width=0.56\linewidth]{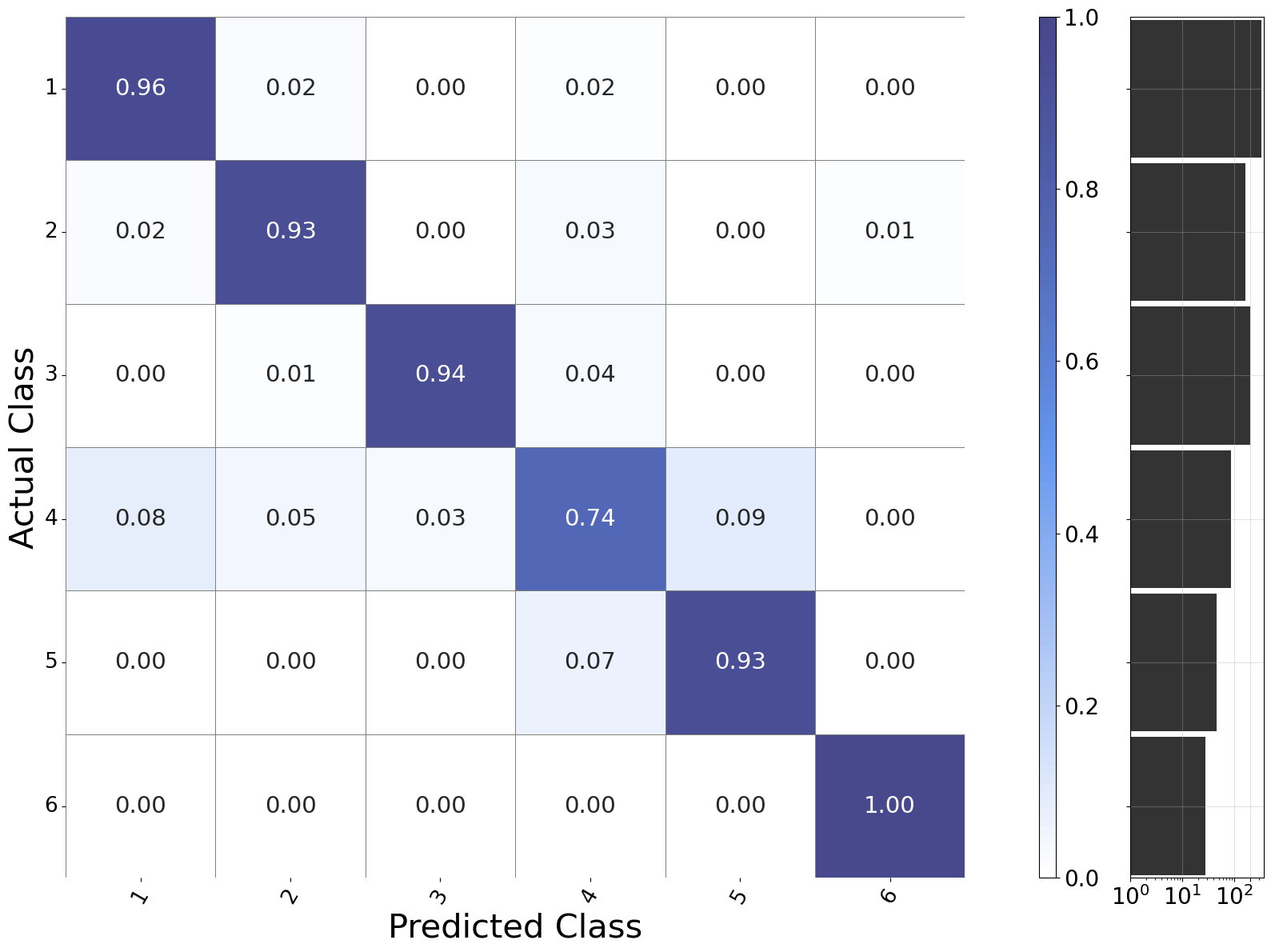}
    \caption{Confusion matrix representing Llama-3.1 70B accuracy in predicting the enzyme class given reactants and substrates, for one experiment. The out-of-diagonal elements show how examples are misclassified. 
    The histogram on the right shows the test set distribution stratified by main class, roughly following how training data is distributed.}
    \label{fig:ec_conf_matr}
\end{figure}

\paragraph{EC prediction task}\mbox{}\\
The 70B model accuracy for EC class prediction is consistent across most classes, with an average accuracy of \textbf{91.7\%}. 
This indicates that it is fairly simple for the fine-tuned model to correctly assign the highest EC number given any reactant, product pair as request.
However, class $4$ exhibits a noticeable performance dip, despite not being the least-represented class in the dataset. To explore the model's misclassification patterns, we present the confusion matrix for EC class prediction in Figure \ref{fig:ec_conf_matr}. 
The matrix reveals that classes \textit{4} and \textit{5} are sometimes wrongly assigned to each other. In classes \textit{1}, \textit{2} and \textit{3} rare instances of misclassifications either happen between \textit{1} and \textit{2} or assign the reactions to class \textit{4}.

\begin{figure}[h!]
    \centering
    \includegraphics[width=0.7\linewidth]{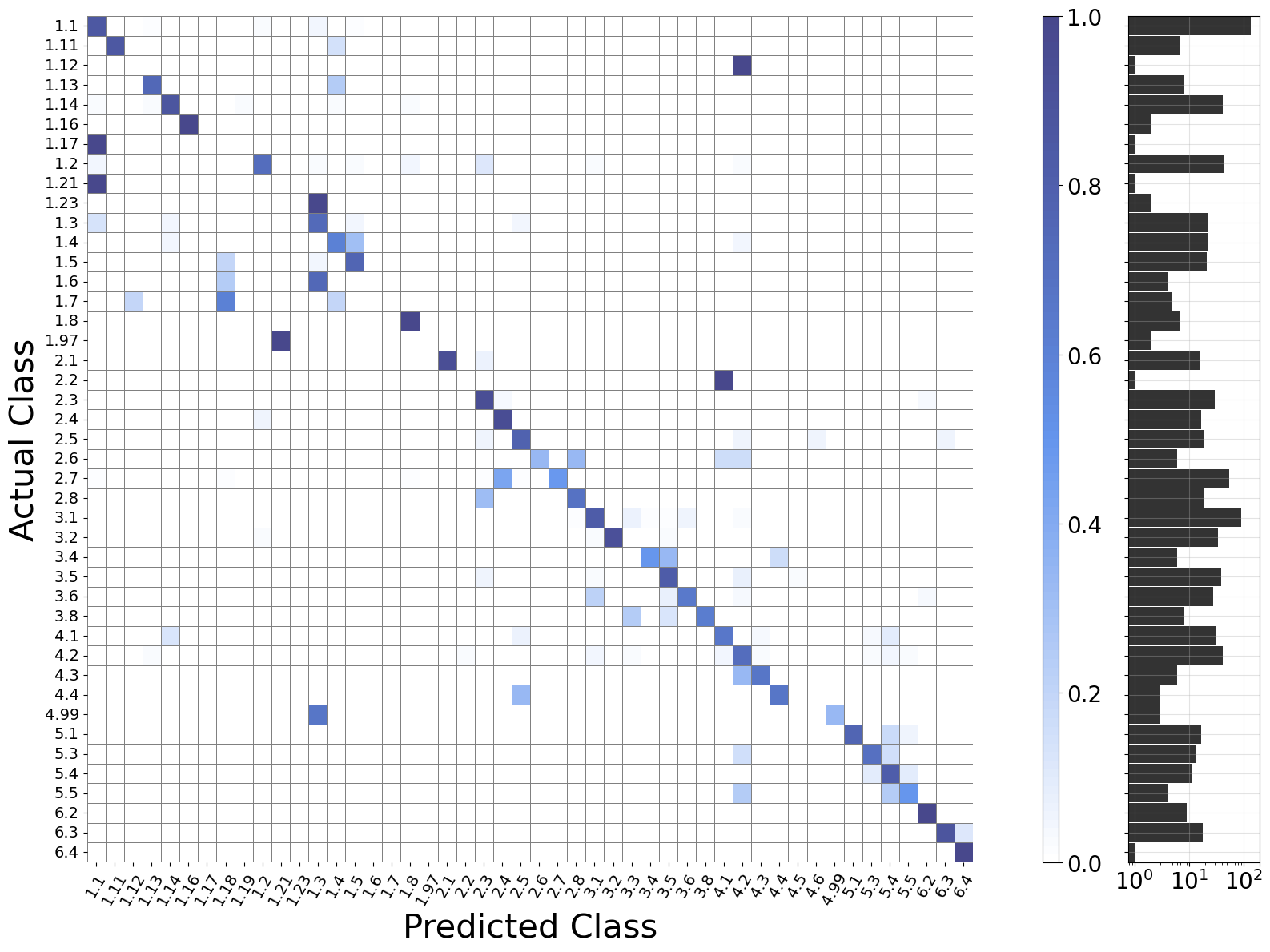}
    \caption{Confusion matrix representing Llama-3.1 70B accuracy in predicting the EC number up to the second digit (EC2), given reactants and substrates, for one experiment. The out-of-diagonal elements show how examples are misclassified. 
    The misalignment in the diagonal elements is due to the set of predicted classes having elements that are not present in the test set, like subclasses $3.3.X.X$ and $4.5.X.X$, that the model predicts in a few cases.
    The histogram on the right shows the test set distribution stratified by EC2 subclass.}
    \label{fig:conf_matr_EC2}
\end{figure}

For EC2 predictions, we see that the model frequently misclassifies subclasses within the same main class. This relates to the EC2 category distribution per main class. For instance, class $1.X.X.X$ has numerous subclasses, whereas classes $5.X.X.X$ and $6.X.X.X$ only have a few. Rare subclasses, such as $2.2.X.X$ or $4.99.X.X$, show clear exceptions with the model misclassifying outside the main class, likely due to their underrepresentation. Additionally, structural similarities within main classes may further contribute to confusion, independent of dataset imbalance. The confusion matrix for the EC2 level, alongside the test set distribution for that depth, is shown in Figure \ref{fig:conf_matr_EC2}. 

\begin{figure}[h!]
    \centering
    \includegraphics[width=0.61\linewidth]{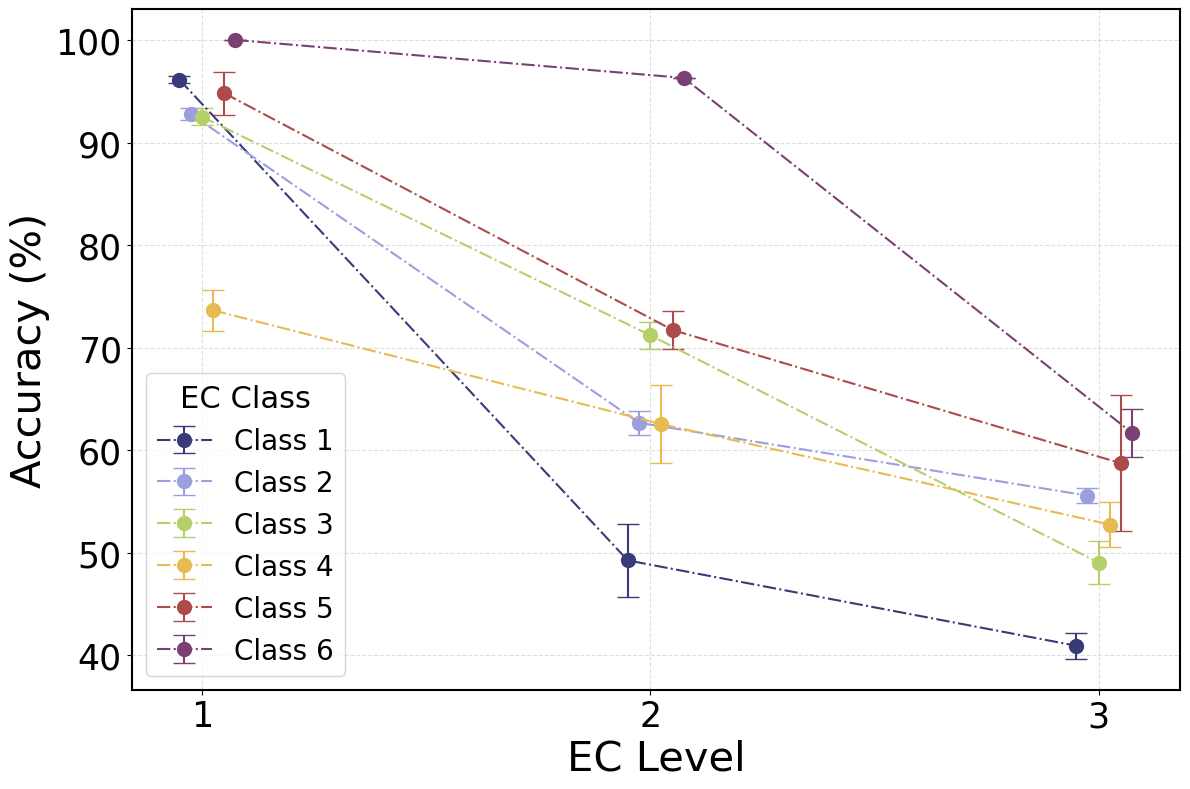}
    \caption{Llama-3.1 70B accuracy in predicting the EC number up to level EC3, organized by main class. Accuracy measures if the model correctly matches the ground truth EC number up to the EC level specified on the $x$-axis. 
    Accuracies are computed considering each (sub)class as equally weighted. These distribution patterns influence model performance, irrespective of reaction complexity or SMILES grammar.}
    \label{fig:acc_vs_level_strat}
\end{figure}

The accuracy of the model declines at deeper EC levels, reflecting the increasing challenge of capturing hierarchical enzyme relationships. These difficulties also stem from increased combinatorial complexity of sublevels and class imbalance. In fact, at level EC2 the model performs best for class \textit{6} and worst for class \textit{1}, a result that aligns with the dataset distribution shown in Figure \ref{fig:brenda_pie}: EC class \textit{1} has a highly branched EC2 structure, with $1.1.X.X$ accounting for almost half of the samples, introducing class imbalance. Conversely, class \textit{6} has a limited number of balanced subcategories ($6.2.X.X$ and $6.3.X.X$), simplifying subclass predictions. Figure \ref{fig:acc_vs_level_strat} illustrates the model's performance in predicting EC numbers up to level EC3, stratified by main class. 

The fine-tuned 70B model comes on top of the fine-tuned 8B model predicting EC digits at any depth. However, the compared SOTA retains a significant edge across all levels (EC1 accuracy: $96.2\%$, EC2 accuracy: $93.4.6\%$, EC3 accuracy: $91.6\%$) \cite{qianGeneralModelPredicting2024}. Please note that the authors have performed a micro-average, while we perform a macro-average that takes class imbalance into account. Extended metrics (F1 score, precision, recall) for the EC class prediction task are reported in Appendix Figure \ref{Appendix:radar_plots}. Additionally, we compare our fine-tuned models with a zero-shot baseline with Llama-3.1 70B, in Table \ref{tab:ec_prediction_results}. We see that the zero-shot prompting approach lacks far behind the fine-tuned models of this size and general capabilities. This indicates that at present it seems inevitable to fine-tune the general purpose model for a complex and domain-specific task such as EC classification in biochemistry.

\begin{table}[h!]
\centering
\renewcommand{\arraystretch}{1.3}
\caption{Performance comparison between Llama-3.1 70B and Llama-3.1 8B models fine-tuned for the EC Prediction task, from predicting level EC1 only, to all digits up to EC3 included. A baseline 0-shot prompting approach with the 70B model is reported as well. We also show our models performance in micro-average next to the SOTA model \cite{qianGeneralModelPredicting2024} in micro-average. Note that the dataset is not exactly the same (see Subsection \ref{sec:task_and_dataset}) and thus results are still not entirely comparable.}

\label{tab:ec_prediction_results}
\begin{tabular}{l|c c c| c c}
\thickhline
\textbf{Metric}          & \textbf{Llama 8B} & \textbf{Llama 70B} & \textbf{Llama 70B 0-shot} & \textbf{Llama 70B micro-avg}& \textbf{SOTA}\\ 
EC1 Accuracy (\%)          & 86.4$\pm$0.6              & \textbf{91.7$\pm$ 0.5}         & 29.6$\pm$ 0.7    &  92.4$\pm$0.2  &  96.2   \\ 
EC2 Accuracy (\%)          & 56.5$\pm$1.5              & \textbf{61.7$\pm$1.1}         & 8.7$\pm$ 0.5      &   75.6$\pm$0.1 &  93.4   \\ 
EC3 Accuracy (\%)          & 40.5$\pm$0.6              & \textbf{49.2$\pm$ 0.7}         & 5.7$\pm$ 0.4     &  68.1$\pm$0.1  &  91.6   \\ 
Validity (\%)          & >99.9              & \textbf{100.0}         & 89.4$\pm$0.3      &  100.0  & - \\ \thickhline
\end{tabular}
\end{table}

\begin{figure}[h!]
    \centering
    \begin{subfigure}[h!]{0.49\textwidth}
        \centering
        \includegraphics[width=\textwidth]{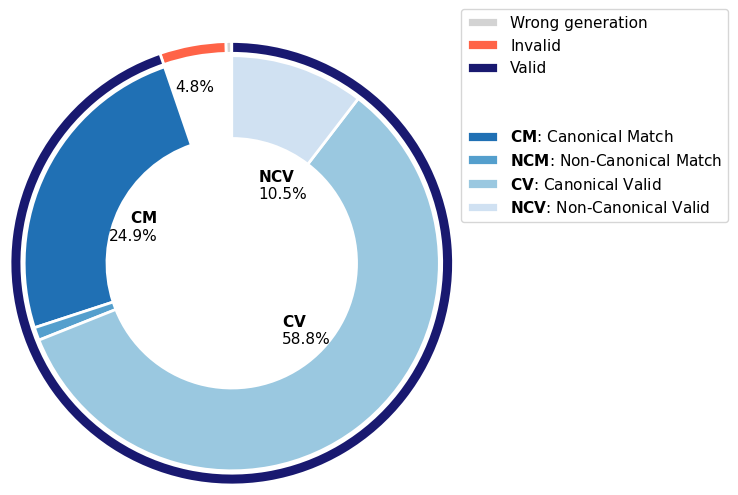}
    \end{subfigure}
    \begin{subfigure}[h!]{0.49\textwidth}
        \centering
        \includegraphics[width=\textwidth]{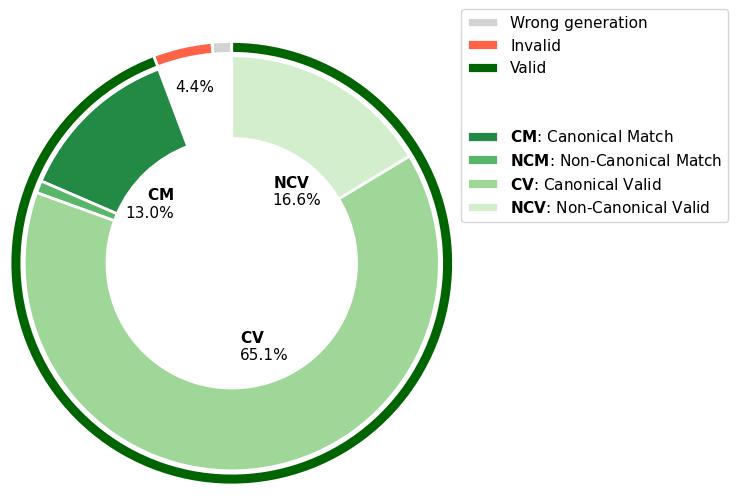}
    \end{subfigure}
    \caption{Pie charts showing the average distribution of predictions for forward synthesis (FS, left) and retrosynthesis (RS, right) for Llama-3.1 70B. The outer layer indicates the proportion of correctly generated (blue/green), invalid chemicals (red), and wrongly generated predictions (grey), while the inner layer differentiates correct outputs from structurally valid but incorrect outputs. 
    Invalid and wrongly formatted predictions remain $<5\%$ and $<2\%$ for both tasks, respectively. Results for each category are obtained averaging over $N=3$ experiments, with standard deviations below $5\%$ of each category value. Percentages are shown for $>2\%$ slices only.}
    \label{fig:pies}
\end{figure}

\paragraph{Product and substrate prediction tasks}\mbox{}\\
The 70B model generates a high proportion of chemically valid molecules in canonical format, with canonical matches (the output string matches the ground truth string as it is) accounting for \textbf{24.9\%} and \textbf{13.0\%} for products and substrates respectively. While FS shows a higher percentage of canonical matches, RS has a greater proportion of chemically valid but incorrect predictions, indicating that retrosynthesis may involve more complex structural reasoning. Chemically invalid predictions are minimal ($<5\%$ of the total test set for both tasks), and wrong generations due to \textit{e.g.} formatting errors are rare ($<2\%$). This demonstrates that the LLMs can easily adhere to complex domain specific grammar like SMILES and to requested output formats, which is a useful property for the analysis of model results. However, these results are not yet competitive with the SOTA model \cite{probstBiocatalysedSynthesisPlanning2022} (49.6\% and 60.0 \% accuracy for exact matches for FS and RS respectively). Note that the dataset used for our models is not exactly the same as the one from SOTA, making the results not directly comparable. Pie charts in Figure \ref{fig:pies} display the distribution of predictions across the five categories for FS and RS tasks, respectively, for Llama-3.1 70B.

\begin{figure}[htbp]
    \centering
    \begin{subfigure}[t]{0.49\textwidth}
        \centering
        \includegraphics[width=\textwidth]{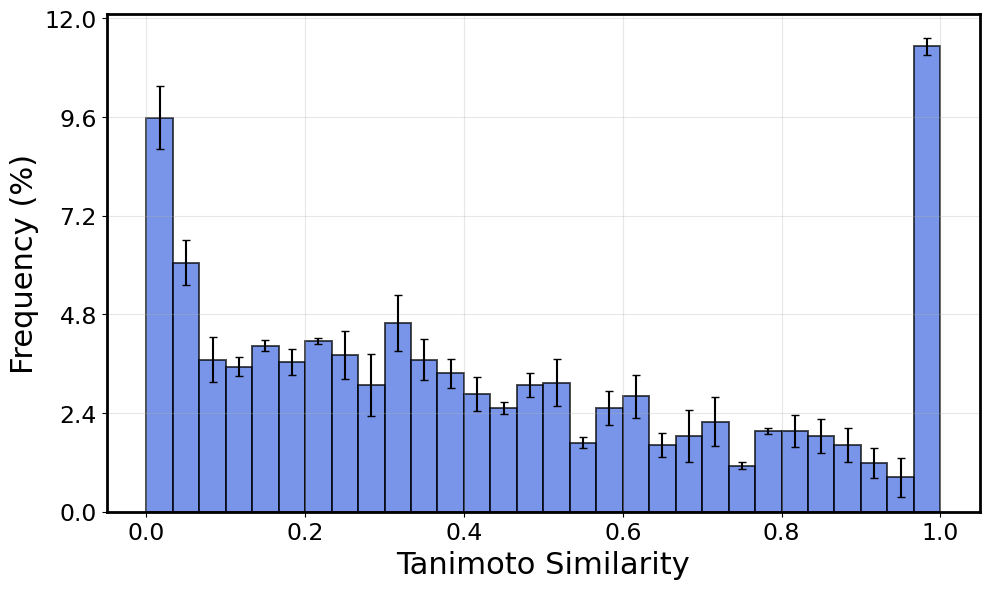}
    \end{subfigure}
    \begin{subfigure}[t]{0.49\textwidth}
        \centering
        \includegraphics[width=\textwidth]{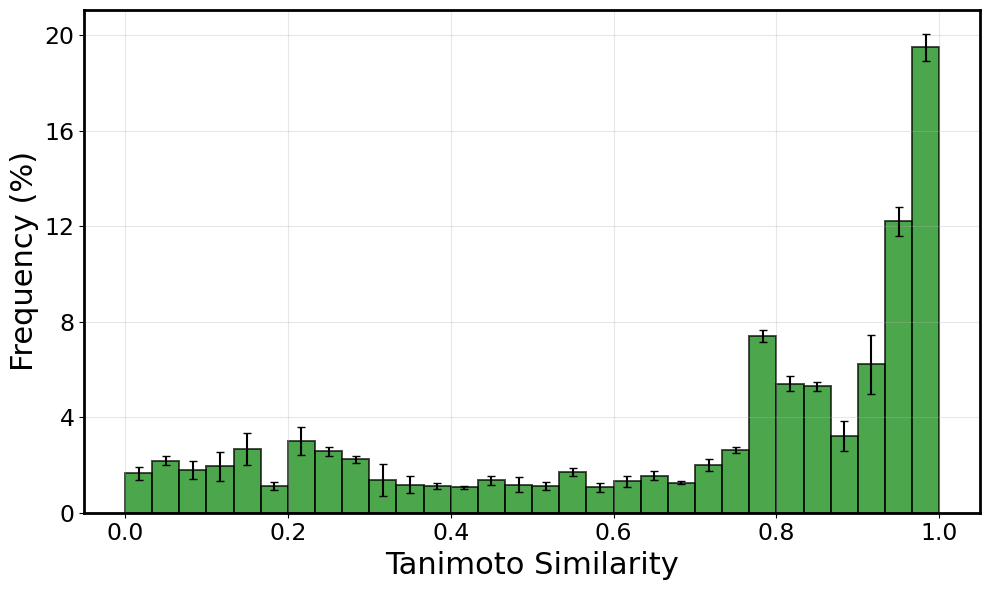}
    \end{subfigure}
    \caption{Histograms of Tanimoto similarities of ground truths against products (\textit{left}) and substrates (\textit{right}), that the model predicts as chemically possible but not corresponding to the ground truth.}
    \label{fig:tanimoto}
\end{figure}

When the model fails to predict the exact molecule, it generates relevant alternatives that may hold biochemical utility in 12\% and 35\% of the cases, for products and substrates respectively. We classify such an output with biochemical utility if the generated molecule shows a high Tanimoto similarity to the correct output. Focusing on the set of valid chemicals, Tanimoto similarity scores are computed and shown in Figure \ref{fig:tanimoto}. In the dataset, SMILES for products are shorter than substrates on average, and we also observe that for branching reactions, the set of products that are possible from certain substrates in a forward synthesis task, is generally smaller than the set of possible substrates reachable from a product in a retrosynthesis task, as observed in the Appendix Figure \ref{fig:count_branches}. Thus, for products, the model either predicts a molecule very close to matching the ground truth, or it gets the wrong chemical. For substrates on the other hand, having longer strings and more options in the RS task leads to generating many substrates that are not correct, but show a relatively high Tanimoto score.
Analyzing the highest Tanimoto values, we see that \textbf{7.1\%} of chemically valid products, and \textbf{6.7\%} of chemically valid substrates, report a score equal to 1. Examples of these chemicals are reported in Appendix Figures \ref{Appendix:smiles_product} and \ref{Appendix:smiles_substrate}. 
We summarize the results for both Llama-3.1 8B and Llama-3.1 70B on FS and RS tasks, including a baseline 0-shot performance and comparison to SOTA in the Appendix Table \ref{Appendix:substrate_product_results}.

\paragraph{Generalization over unseen tasks}\mbox{}\\
When fine-tuned on a single biochemical task, the model not only retains its general capabilities on unseen, related tasks within the same sub-domain but also improves its performance compared to its zero-shot baseline. To evaluate this generalization effect, we test each of the three single-task (ST) fine-tuned models on the two tasks they were not trained on, comparing their performance to the respective zero-shot baseline.
Results show that fine-tuning on either the FS or RS task significantly improves EC class prediction accuracy, nearly doubling the zero-shot baseline performance. Likewise, a model fine-tuned exclusively on EC number prediction improves FS match accuracy from nearly 0\% to 12.9\% while also reducing invalid predictions by half.
Table \ref{tab:generalize} presents the generalization results, where each fine-tuned model is tested on the two unseen tasks.

\begin{table}[h!]
\centering
\renewcommand{\arraystretch}{1.3}
\caption{Generalization of ST fine-tuned Llama-3.1 70B models when tested on the unseen related biochemical tasks. The zero-shot baseline is reported for comparison. Performance on the original fine-tuned task is omitted to emphasize cross-task generalization. The reported \textit{Match} values are here considered regardless of canonicity. The \textit{Invalid} category includes both incorrect SMILES notation as well as wrongly formatted output from the LLM.}
\label{tab:generalize}
\resizebox{\textwidth}{!}{ 

\begin{tabular}{c cc cc cc}
\thickhline
\multirow{2}{*}{\textbf{Fine-tuned on}} 
    & \multicolumn{2}{c}{\textbf{EC}} 
    & \multicolumn{2}{c}{\textbf{FS}} 
    & \multicolumn{2}{c}{\textbf{RS}}\\ 
\cmidrule(lr){2-3} \cmidrule(lr){4-5} \cmidrule(lr){6-7}
 & EC1$\uparrow$ (\%) & Invalid$\downarrow$ (\%) & Match$\uparrow$ (\%) & Invalid$\downarrow$ (\%) & Match$\uparrow$ (\%) & Invalid$\downarrow$ (\%) \\ 
\midrule
\textbf{EC} & - & - & \textbf{12.9} & 31.1 & 0.3 & 55.3 \\ 
\textbf{FS} & \textbf{54.3} & \textbf{0.3}& - & - & \textbf{1.4} & \textbf{3.3} \\ 
\textbf{RS} & 42.1 & 6.3 & 0.6 & \textbf{5.5} & - & - \\ 
\midrule
\textbf{ICL 0-shot} & 29.6 & 10.6 &<0.1 & 53.5 & <0.1& 82.4\\ 
\thickhline
\end{tabular}

} 
\end{table}

\subsection{Multitask fine-tuning}

Using a multitask setup we show that we can improve performance through the use of synergistic information from the related task, in particular for FS and RS tasks. For these the model performance for matches (regardless of canonicity) increases by \textbf{7.9\%} and \textbf{5.3\%} respectively.
The three ST datasets are merged together to provide the dataset used for the MT setup. The Llama 70B and 8B models are both fine-tuned, using the best-performing configuration identified in the single-task experiments. Performance is compared against single-task setups to assess multitask learning benefits, with the main results reported in Table \ref{tab:st_vs_mt_results}.

\begin{table}[h!]
\centering
\renewcommand{\arraystretch}{1.3}
\caption{Performance comparison between single-task and multitask setups for Llama-3.1 8B and Llama-3.1 70B. Blue cells represent performance improvement, orange cells represent performance reduction. The reported \textit{Match} values are here considered regardless of canonicity. The categories "\textit{Match+(TS=1)}" and "\textit{Match+(TS>0.95)}" add to the previous one the share of valid chemicals with a Tanimoto score equal to 1 and greater than 0.95 respectively.
Numbers are presented in bold if the best performance improvement does not fall within one standard deviation from the second-best.}
\label{tab:st_vs_mt_results}
\begin{tabular}{l l c c c}
\thickhline
\textbf{Task} & \textbf{Metric (\%)} & \multicolumn{3}{c}{\textbf{Llama-3.1 70B}} \\ \hline
              &                 & \textbf{ST} & \textbf{MT} & \textbf{$\Delta$}  \\ \hline
\multirow{2}{*}{\textbf{EC}} 
              & Accuracy EC1 $\uparrow$  & \textbf{91.7}        & 86.4        & \cellcolor{orange!40} -5.3           \\ 
              & Accuracy EC2 $\uparrow$  & 61.7        & \textbf{65.1}        & \cellcolor{blue!25} +3.7           
              \\
              & Accuracy EC3 $\uparrow$  & 49.2        & 48.9        & -0.3           \\ \narrowhline
\multirow{3}{*}{\textbf{FS}} 
              & Match $\uparrow$& 25.9        & \textbf{33.8}        & \cellcolor{blue!25} +7.9           \\ 
              & Match+(TS=1) $\uparrow$  & 33.0        &     \textbf{44.4}    & \cellcolor{blue!25} +11.4           \\ 
              & Match+(TS>0.95) $\uparrow$  & 34.2        &     \textbf{45.4}    & \cellcolor{blue!25} +11.2           \\ 
              & Invalid $\downarrow$         & 4.8        & 4.9        & +0.1          \\ \narrowhline
\multirow{3}{*}{\textbf{RS}} 
              & Match $\uparrow$& 13.9        & \textbf{19.2}        & \cellcolor{blue!25} +5.3          \\ 
              & Match+(TS=1) $\uparrow$   & 20.6        &    \textbf{30.1}     & \cellcolor{blue!25}  +9.5          \\ 
              & Match+(TS>0.95) $\uparrow$   & 36.1        &    \textbf{45.4}     & \cellcolor{blue!25}  +9.3          \\ 
              & Invalid $\downarrow$        & 4.4        & \textbf{3.0}        & \cellcolor{blue!25} -1.4          \\ \thickhline
\end{tabular}
\end{table}

\subsection{Exploring low-data regimes}

Fine-tuned LLMs show promise in low data regimes: for Llama-3.1 70B, we report almost double EC class accuracy when comparing zero-shot prompting (29.6\%) with the fine-tuned version with only N=200 training samples (55.3\%). We replicate low-data scenarios to evaluate how the models perform with significantly reduced training samples. Specifically, we analyze performance degradation when the training set size is limited to 600 and 200 compared to our default training ($\sim$1800 samples per task). This analysis is conducted for both models, to provide insights into their scalability when data availability becomes the bottleneck. Both models show a steady performance increase when training data is increased. The larger architecture holds an edge over the smaller one regardless of data size across almost all tasks, confirming again its greater capabilities. 

For a fairer comparison, we include a simple XGBoost baseline. XGBoost \cite{chenXGBoostScalableTree2016} is a gradient boosting model that performs well with structured data and does not rely on large-scale pretraining, making it a suitable reference for evaluating whether LLM fine-tuning truly adds value in data-limited biochemical prediction tasks. We find that across all tasks and for each data scenario, our models outperform the XGBoost model. We report our findings in Table \ref{tab:lowdata}. More details on how XGBoost is trained are reported in the Appendix in subsection \ref{app:XGBoost}.

\begin{table}[h!]
\centering
\renewcommand{\arraystretch}{1.3}
\caption{Performance of Llama-3.1 8B and Llama-3.1 70B across all tasks and for different training set sizes. The reported \textit{Match} values are here considered regardless of canonicity. Each task is trained on a slightly different amount of samples ($\pm$ 20) because of how data has been split, thus we report a reference number of 1800 samples in the corresponding rows. Numbers are presented in bold if the best performance does not fall within one standard deviation from the second-best. A baseline XGBoost model is reported for comparison.}
\label{tab:lowdata}
\resizebox{\textwidth}{!}{ 

\begin{tabular}{l c ccc cc cc}
\thickhline
\multirow{2}{*}{\textbf{Model}} & \multirow{2}{*}{\textbf{Train set size}} 
    & \multicolumn{3}{c}{\textbf{EC}} 
    & \multicolumn{2}{c}{\textbf{FS}} 
    & \multicolumn{2}{c}{\textbf{RS}} \\ 
\cmidrule(lr){3-5} \cmidrule(lr){6-7} \cmidrule(lr){8-9}
 &  & EC1$\uparrow$ (\%) & EC2$\uparrow$ (\%) & EC3$\uparrow$ (\%) & Match$\uparrow$ (\%) & Invalid$\downarrow$ (\%) & Match$\uparrow$ (\%) & Invalid$\downarrow$ (\%) \\ 
\midrule
\multirow{3}{*}{LLama-3.1 8B}  
    & 200  & 43.5 & 15.5 & 8.5 & 2.6 & 4.6 & 0.2 & 11.2 \\ 
    & 600  & 65.6 & 30.1 & 17.4 & 8.3 & 6.4 & 2.8 & 10.2 \\ 
    & $\sim$ 1800  & 86.4 & 56.5 & 40.5 & 18.4 & 9.4 & \textbf{15.1} & \textbf{4.3} \\ 

\midrule
\multirow{3}{*}{LLama-3.1 70B}  
    & 200  & 55.3 & 28.5 & 17.7 & 7.7 & 7.7 & 2.9 & 7.3 \\ 
    & 600  & 73.5 & 45.8 & 33.1 & 11.0 & 4.4 & 4.1 & 7.2 \\ 
    & $\sim$ 1800 & \textbf{91.7} & \textbf{61.7} & \textbf{49.2} & \textbf{25.9} & 4.8 & 13.9 & \textbf{4.4} \\ 
\midrule
\multirow{3}{*}{XGBoost}  
    & 200  & 32.7 & 4.9 & \textless 0.1 & \textless 0.1 & - & \textless 0.1 & - \\ 
    & 600  & 40.9 & 6.0 & 1.7 & 1.9 & - & 2.5 & - \\ 
    & $\sim$ 1800 & 54.0 & 23.7 & 15.9 & 5.1 & - & 3.6 & - \\ 
\thickhline
\end{tabular}

} 
\end{table}

\subsection{Impact of different LoRA setups}

We observe that adding more trainable parameters can lead to performance improvement for most tasks. This indicates the importance of parameter-efficient learning strategies in domains where fine-tuning is essential. We see the trend that LoRA default performs better than LoRA attention and LoRA light in almost all settings. In most tested cases for the 8B model, LoRA attention performs slightly better than LoRA light, while for the 70B model, LoRA light performs slightly better than LoRA attention in most tested cases. Performance across all tasks with different LoRA setups are reported in Table \ref{tab:multiple_loras}.

\begin{table}[h!]
\centering
\renewcommand{\arraystretch}{1.3}
\caption{Performance of Llama-3.1 8B and Llama-3.1 70B across all tasks and for different fine-tuning setups. Perfomance  for all tasks increases with the number of fine-tuned parameters, the only exception being the \textit{attention} fine-tuning for Llama-3.1 70B, where an increase in FS performance comes with a degradation in RS and EC prediction tasks. The reported \textit{Match} values are here considered regardless of canonicity. Numbers are presented in bold if the best performance does not fall within one standard deviation from the second-best.} 
\label{tab:multiple_loras}
\resizebox{\textwidth}{!}{ 
\begin{tabular}{l c ccc cc cc}
\thickhline
\multirow{2}{*}{\textbf{Model}} & \multirow{2}{*}{\textbf{LoRA type}} 
    & \multicolumn{3}{c}{\textbf{EC}} 
    & \multicolumn{2}{c}{\textbf{FS}} 
    & \multicolumn{2}{c}{\textbf{RS}} \\ 
\cmidrule(lr){3-5} \cmidrule(lr){6-7} \cmidrule(lr){8-9}
 &  & EC1$\uparrow$ (\%) & EC2$\uparrow$ (\%) & EC3$\uparrow$ (\%) & Match$\uparrow$ (\%) & Invalid$\downarrow$ (\%) & Match$\uparrow$ (\%) & Invalid$\downarrow$ (\%) \\ 
\midrule

\multirow{3}{*}{LLama-3.1 8B}  
    & light  & 72.0 & 44.1 & 30.3 & 10.2 & 9.1 & 4.7 & 12.8 \\ 
    & attention  & 82.0 & 48.4 & 31.9 & 11.3 & 7.9 & 6.9 & 10.1 \\ 
    & default & 86.4 & 56.5 & 40.5 & 18.4 & 9.4 & \textbf{15.1} & 4.3 \\ 

\midrule

\multirow{3}{*}{LLama-3.1 70B}  
    & light  & 85.8 & 58.5 & 45.2 & 21.4& 6.0 & 13.7 & 3.9 \\ 
    & attention  & 78.8 & 48.0 & 34.9 & \textbf{25.6}& 5.5 & 9.8 & 3.3 \\ 
    & default & \textbf{91.7} & \textbf{61.7} & \textbf{49.2} & \textbf{25.9} & 4.8 & 13.9 & 4.4 \\ 

\thickhline
\end{tabular}

} 
\end{table}

\subsection{Limitations}

While our study demonstrates the potential for researchers to work with LLMs when studying biochemical reactions, several limitations must be acknowledged. Addressing these will be key to improving both model accuracy and applicability in real-world biochemical workflows.

\begin{itemize}
    \item Potential data leakage: although we fine-tune the LLM to evaluate performance in low-data regimes, it is possible that the model has already been exposed to similar biochemical reaction data during pretraining, as such datasets are available online. For a fairer comparison, future evaluations should ensure that test sets are composed of truly held-out reactions that cannot be scraped or indirectly inferred from pretraining text on the internet. This would provide a clearer measure of the model's generalization ability beyond memorization.
    \item Data constraints: our study is based on the BRENDA subset of the ECREACT dataset, which, while extensive, does not fully cover the diversity of enzymatic reactions and does not allow a direct comparison to current SOTA model. The limited representation of certain EC subclasses affects generalization. Expanding training to the full ECREACT dataset or integrating additional reaction databases could mitigate this issue and enhance model robustness, yet also here, ECREACT has been preprocessed and simplifies complex biochemical reaction mechanisms to a certain degree.
    \item Computational constraints: fine-tuning LLMs is computationally expensive, even with PEFT strategies like LoRA, limiting accessibility for resource-constrained environments. 
\end{itemize}

\section{Conclusions}

In this study, we systematically evaluated the potential of Large Language Models (LLMs) for biochemical reaction prediction, focusing on Enzyme Commission classification, forward synthesis, and retrosynthesis. By fine-tuning Llama-3.1 models, we demonstrated that LLMs can answer biochemical questions, although they are not yet fully competitive with specialized models. Fine-tuning significantly improves performance over in-context learning, with Llama-3.1 70B achieving 91.7\% accuracy in EC class classification. Fine-tuning on a single task does not degrade the 70B model capabilities on unseen related tasks, as we observe performance improvement compared to zero-shot baselines that use the base, pretrained model. Multitask learning enhances forward synthesis and retrosynthesis predictions, with a match accuracy of 33.8\% and 19.2\% respectively, indicating that leveraging shared biochemical knowledge improves generalization. Additionally, LLMs have potential in low-data regimes, making them valuable for applications where labeled data is scarce. The choice of fine-tuning strategy impacts the performance, with LoRA offering an efficient and scalable adaptation method.
Despite these strengths, several challenges remain: LLMs struggle with handling rare EC subclasses and ensuring reliable predictions. As LLM architectures continue to evolve, their integration into biochemical workflows has the potential to accelerate discoveries in enzyme-substrate prediction and biocatalysis design.

\subsection{Data availability statement}
Data and code for this article are available at \url{https://github.com/Intelligent-molecular-systems/LLM_finetuning_for_biochemistry}. This study was carried out using publicly available data at \url{https://github.com/rxn4chemistry/biocatalysis-model}.

\subsection{Author Contributions}
J. M. Weber and L. Di Fruscia conceptualized the study. L. Di Fruscia led data curation, formal analysis, investigation, methodology, software development, validation, visualization, and drafting of the manuscript. J. M. Weber supervised the project, provided resources, and reviewed and edited the manuscript.

\subsection{Conflict of interest}
There are no conflicts to declare.
\label{sec:others}

\bibliographystyle{unsrt}  
\bibliography{references}  

\newpage
\section{Appendix}
\setcounter{table}{0}
\renewcommand{\thetable}{A\arabic{table}}
\setcounter{figure}{0}
\renewcommand{\thefigure}{A\arabic{figure}}

\subsection{Data preprocessing and analysis}
\label{Appendix:preprocessing}
We implemented a series of preprocessing steps to ensure a fair split across training and test set and across tasks: 
\begin{itemize}
    \item Canonicalization of SMILES representations: reactions with substrates or products in different SMILES representations are unified by converting all SMILES strings to their canonical forms. This ensures that duplicate $\{substrate, product\}$ pairs, differing only in molecular representation, are identified and removed.
    \item Grouping of related reactions: reactions that represent the same underlying biochemical process but differ slightly due to variations in substrate or product representations (if there are duplicate $\{product, EC\}$ or $\{substrate, EC\}$ pairs) are grouped. We refer to this as $\textit{substrate branching}$ and $\textit{product branching}$ respectively. All reactions within a group are allocated to the same dataset split (training or test) to avoid leakage. 
    \item Avoidance of task-specific leakage: in forward synthesis (FS) and retrosynthesis (RS), if a reaction appears in FS, then any of its counterparts with the same product and EC number but different substrates, must not appear in RS. This prevents the model from gaining undue advantage by being exposed to related information in the training phase.
\end{itemize}

\begin{figure}[h!]
    \centering
    \includegraphics[width=1.\linewidth]{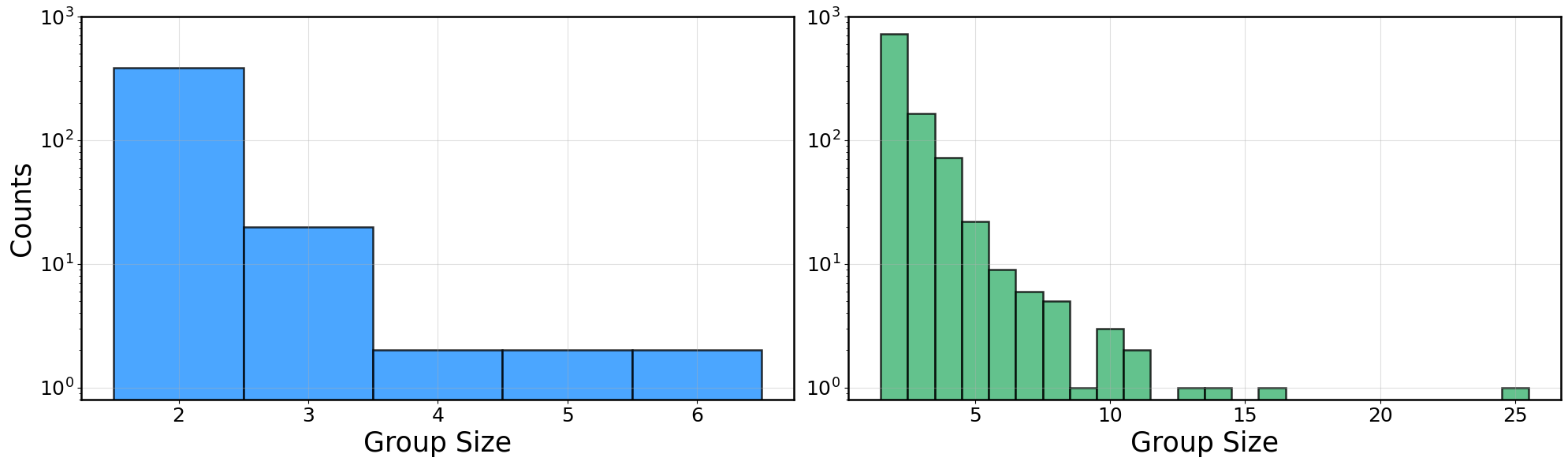}
    \caption{Histograms of group size for duplicate \{\textit{substrate, EC}\} (left) and duplicate \{\textit{product, EC}\} (right). Having a duplicate \{\textit{substrate, EC}\} group of size $N$ implies that, if the group of reactions is used for a product prediction task, that single input branches into $N$ possible outputs. The same reasoning holds for duplicate \{\textit{product, EC}\} involved in a substrate prediction task. We can observe that while most duplicate reactions branch into two possible products, substrates tend to branch into larger groups.}
    \label{fig:count_branches}
\end{figure}

Branching groups distribution varies a lot on whether we look at the products or the substrates, as most of the branching substrates only lead to 2 or 3 possible products, but the reverse task has a wider spread. We report this in Figure \ref{fig:count_branches}. We further analyze the substrates and products to assess whether overlapping reactions across groups are present. This is needed because \textit{e.g.} a substrate, while branching into multiple products, may also be part of a set of substrates reachable from a specific product. We follow this by merging those overlapping groups together and removing redundant entries. 

\newpage
\subsection{EC class prediction radar plots}

\begin{figure}[htbp]
    \centering
    \begin{subfigure}[t]{0.49\textwidth}
        \centering
        \includegraphics[width=\textwidth]{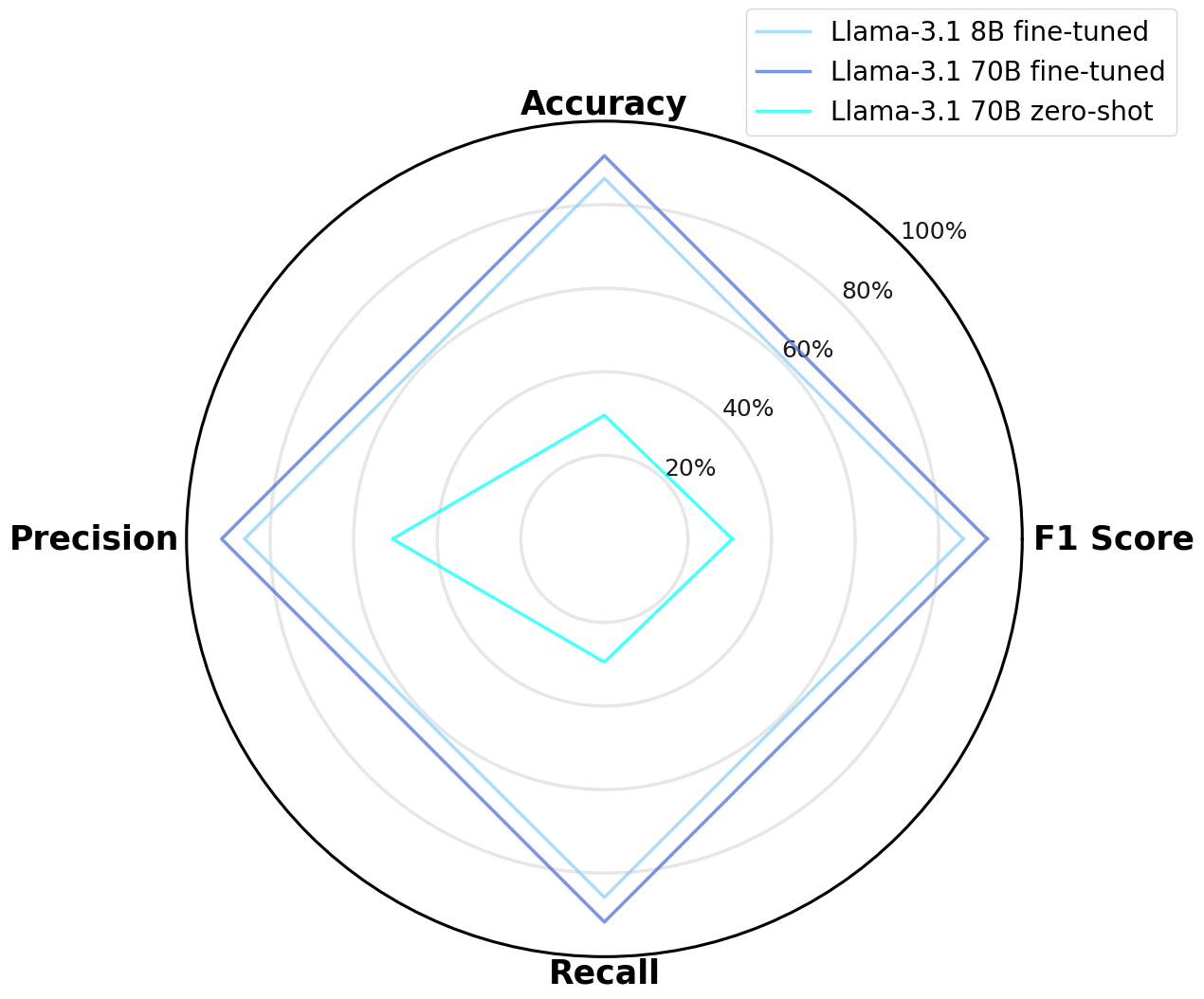}
    \end{subfigure}
    \hspace{0.01\textwidth} 
    \begin{subfigure}[t]{0.49\textwidth}
        \centering
        \includegraphics[width=\textwidth]{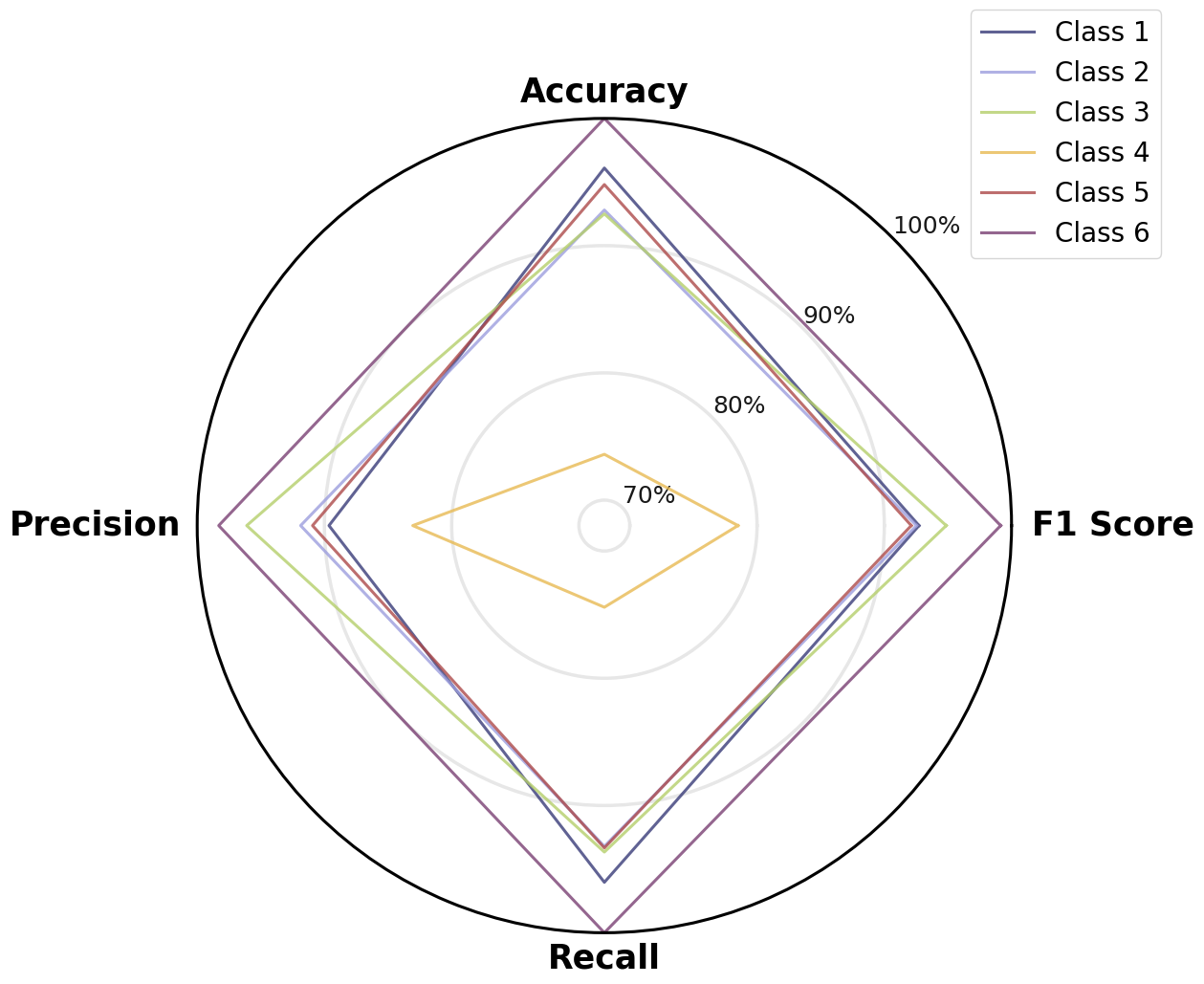}
    \end{subfigure}
    \caption{\textit{Left}: Radar plot showing accuracy, precision, recall and F1 score for Llama-3.1 70B and Llama-3.1 8B averaged over $N=3$ experiments. The plot demonstrates consistent outperformance of the larger model over the 8B variant across all metrics. For comparison, we also show the 70B model performance when it is not fine-tuned, in a zero-shot format. \textit{Right}: EC class accuracy for the fine-tuned Llama-3.1 70B stratified by the class.}
    \label{Appendix:radar_plots}
\end{figure}

Computing Precision, Recall and F1 score alongside Accuracy, we observe that these four metrics are all consistent with each other for both of our fine-tuned model sizes, with Llama-3.1 70B beating Llama-3.1 8B in every metric. We compare them to a 0-shot prompting setup with the pretrained Llama-3.1 70B as a baseline, observing the clear performance gap between in-context learning with the larger model, against the fine-tuned 8B version. Focusing on the fine-tuned 70B model, a stratification by main class shows us again that the values for the four metrics are consistent with each other, per class, with class \textit{4} being the most unbalanced. These findings are reported in Figure \ref{Appendix:radar_plots}.

\subsection{Forward- and Retrosyntesis comparison with fine-tuned Llama 8B}

The 70B model performs better than the 8B one for forward synthesis, and are both comparable when it comes to retrosynthesis. We report the main results in Table \ref{Appendix:substrate_product_results}, alongside the SOTA model. 

\begin{table}[h!]
\centering
\renewcommand{\arraystretch}{1.3}
\caption{Performance comparison between Llama-3.1 8B and Llama-3.1 70B models for forward- and retrosynthesis. All values for our fine-tuned models are obtained averaging over $N=3$ experiments, with standard deviations below $5\%$ of each category value. A zero-shot baseline on the pretrained 70B model is reported for comparison. We also report the SOTA model \cite{probstBiocatalysedSynthesisPlanning2022} performance at the end. Note that the dataset is not exactly the same (see Subsection 2.1) and thus results are still not entirely comparable. Numbers are presented in bold if the best performance does not fall within one standard deviation from the second-best. NCM, CV, and NCV categories taken alone do not reflect model improvement, thus we do not bold them.\\
CM: Canonical Matching, NCM: Non-Canonical Matching, CV: Canonical Valid, NCV: Non-Canonical Valid.}
\label{Appendix:substrate_product_results}
\begin{tabular}{l c c c c c c}
\thickhline
\textbf{Model} & \textbf{Task} & \textbf{CM$\uparrow$ (\%)} & \textbf{NCM (\%)} & \textbf{CV (\%)} & \textbf{NCV (\%)} & \textbf{Invalid$\downarrow$ (\%)} \\ \hline
\multirow{2}{*}{Llama-3.1 8B} 
    & FS & 17.6 & 0.8 & 53.8 & 14.0 & 9.4 \\ 
    & RS & 14.0 & 1.1 & 67.8 & 11.7 & 4.3 \\ \hline
\multirow{2}{*}{Llama-3.1 70B} 
    & FS & \textbf{24.9} & 1.0 & 58.8 & 10.5 & \textbf{4.8} \\ 
    & RS & 13.0 & 0.9 & 65.1 & 16.6 & 4.4 \\ \hline
\multirow{2}{*}{Llama-3.1 70B 0-shot} 
    & FS & \textless 0.1 & 0 & 40.5 & 5.9 & 53.5 \\ 
    & RS & 0 & \textless 0.1 & 12.7 & 4.9 & 82.4 \\ \hline
    \multirow{2}{*}{SOTA}
    & FS & 49.6 & - & - & - & - \\ 
    & RS & 60.0 & - & - & - & -\\
\thickhline
\end{tabular}
\end{table}

\newpage
\subsection{Average Tanimoto scores in ground truth branching}

\begin{figure}[h!]
    \centering
    \includegraphics[width=0.9\linewidth]{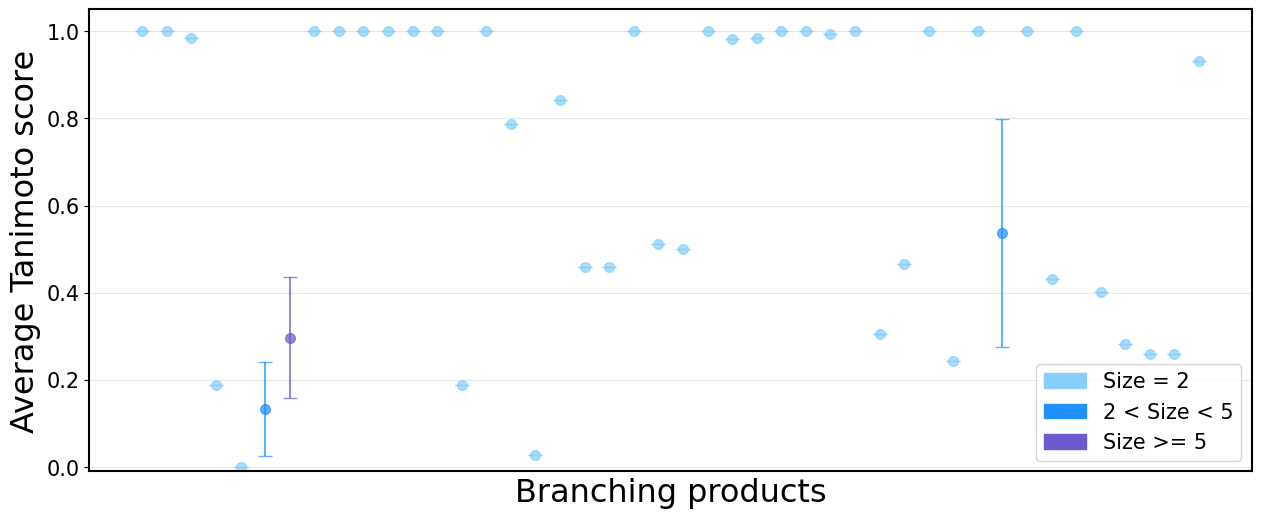}
    \caption{Average Tanimoto score computed across a ground truth product and each of its ground truth branching counterparts, for all groups and stratified by group size. For branching groups of size 2, no standard deviation is shown as we only have one Tanimoto score computed between the reference ground truth and its alternative option.}
    \label{Appendix:avg_tanimoto_prod}
\end{figure}

\begin{figure}[h!]
    \centering
    \includegraphics[width=0.9\linewidth]{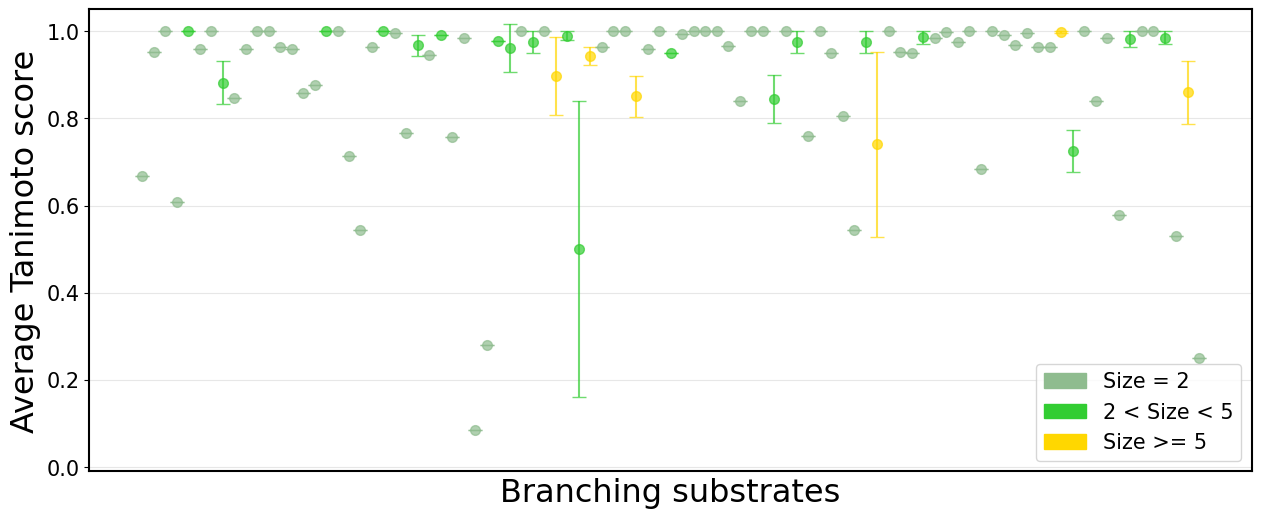}
    \caption{Average Tanimoto score computed across a ground truth substrate and each of its ground truth branching counterparts, for all groups and stratified by group size. For branching groups of size 2, no standard deviation is shown as we only have one Tanimoto score computed between the reference ground truth and its alternative option.}
    \label{Appendix:avg_tanimoto_sub}
\end{figure}

We observe that for the equally valid ground truths that the database stores for a given reaction, many examples show a relatively low similarity score. Focusing on the product prediction only, some of the reasons this happen can be due to having a co-factor recorded in place of the main product, or some entries may report products that correspond to different reaction intermediates, a problem that strictly relates to the presence of branching reactions in the dataset.  
We compute the average Tanimoto score across ground truth chemicals that belong to the same set of branching product/substrates, to get insights over the chemical diversity of alternatives products/substrates that are reported in the dataset.\\
Given a group of size \textit{N}, we compute the Tanimoto scores between one element of the set and the remaining \textit{N-1} chemicals. Then, we compute the average Tanimoto score and its standard deviation for that group. If the chemicals are all similar to each other, we observe a high average with a relatively small standard deviation. On the other end, if the chemicals present more variability, we expect to see a lower average with a wider spread in the standard deviation.
We report the findings in Figures \ref{Appendix:avg_tanimoto_prod} and \ref{Appendix:avg_tanimoto_sub}.

\newpage
\subsection{Predictions with Tanimoto score equal to 1 for products and substrates}
\begin{figure}[h!]
    \centering
    \includegraphics[width=1.\linewidth]{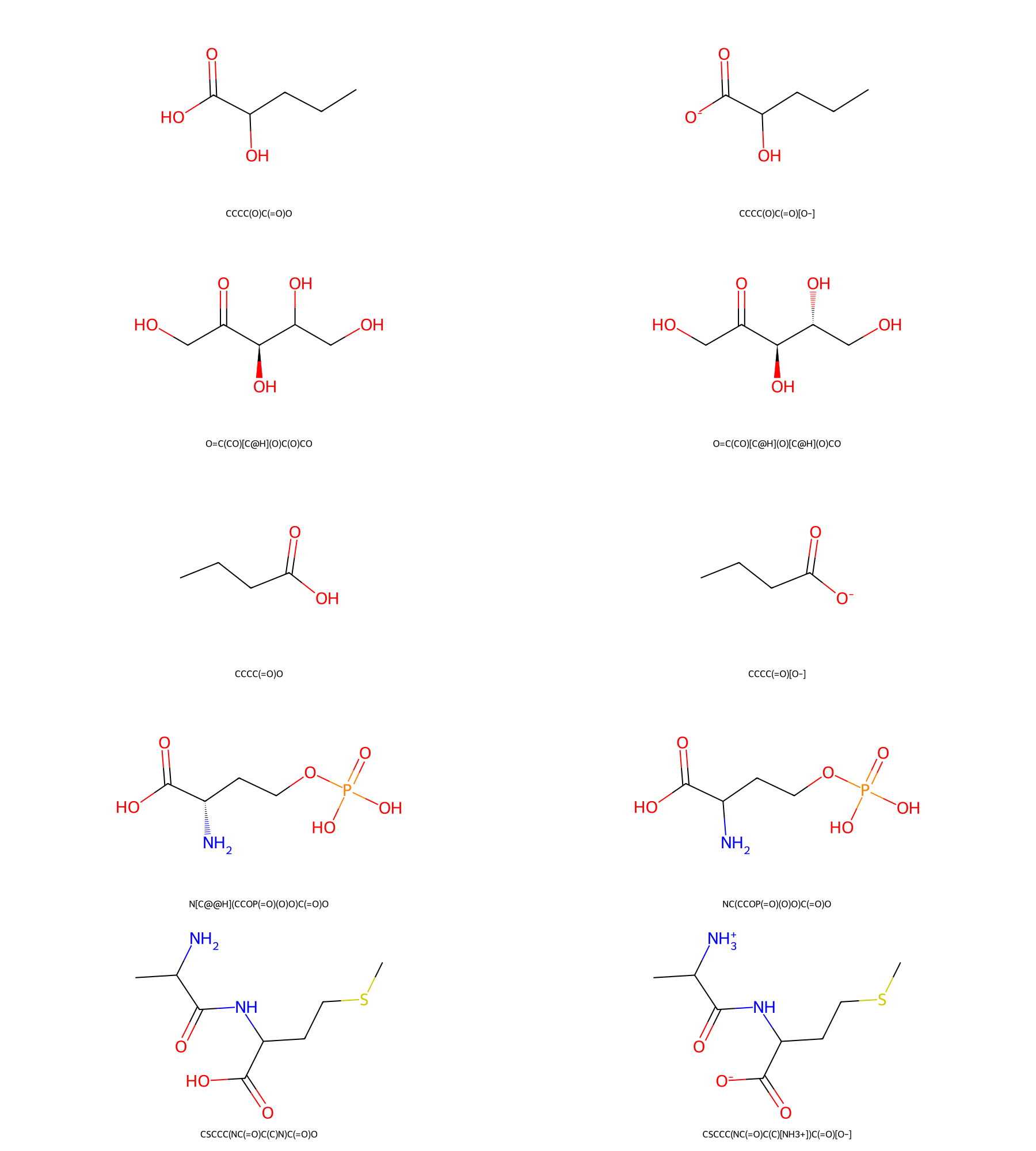}
    \caption{Examples of predicted (\textit{left}) vs ground truth (\textit{right}) products, when the prediction is not correct but produces a Tanimoto score equal to 1. We see that some predictions have an additional hydrogen (resulting in an OH group) while the ground truth recorded an oxygen ion (O-) (rows \textit{1}, \textit{3}), while some others have a mismatch in chirality (rows \textit{2} and \textit{4}).}
    \label{Appendix:smiles_product}
\end{figure}

\newpage

\begin{figure}[h!]
    \centering
    \includegraphics[width=1.\linewidth]{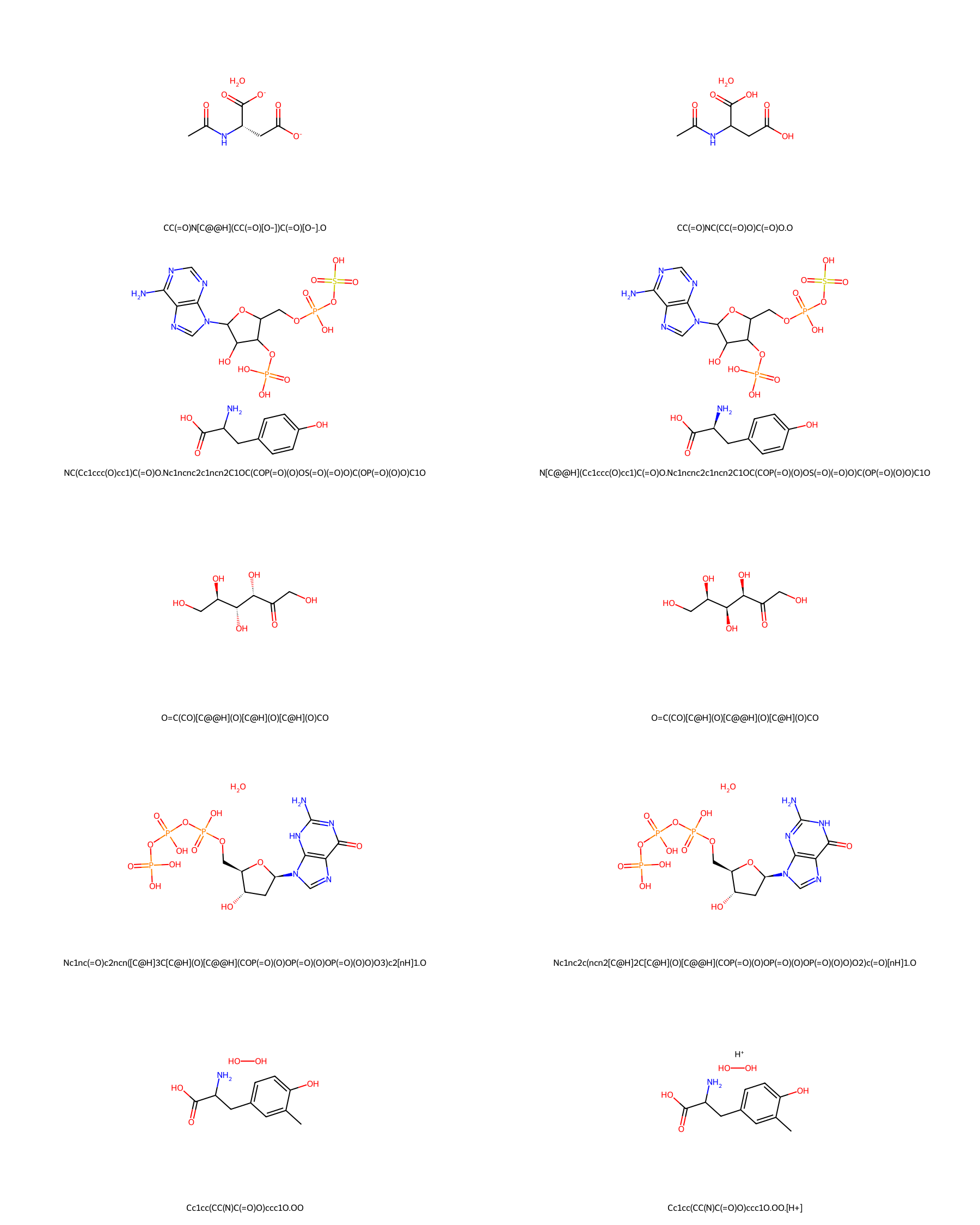}
    \caption{Examples of predicted (\textit{left}) vs ground truth (\textit{right}) substrates, when the prediction is not correct but produces a Tanimoto score equal to 1. We see that some predictions have a missing hydrogen (resulting in an oxygen ion O-) while the ground truth recorded an OH group (row \textit{1}), while some others have a mismatch in chirality (\textit{e.g.} rows \textit{2} and \textit{3}).}
    \label{Appendix:smiles_substrate}
\end{figure}

\newpage
\subsection{XGBoost data preprocessing and training}
\label{app:XGBoost}

For each task, we encode the biochemical inputs into a structured format that XGBoost can process efficiently. Given its reliance on tabular data, molecular and enzymatic information is transformed into numerical feature vectors before being fed into the model:
\begin{itemize}
    \item Molecular representation: for the product and substrate prediction tasks, we represent molecules using Morgan fingerprints to encode molecular structures into a fixed-length binary vector. Each molecule is transformed in a 256-bit binary vector, where each bit represents the presence or absence of a specific chemical substructure.
    \item Reaction representation: for the EC number prediction task, the entire biochemical reaction (substrates + products) is encoded as a 1024-bit reaction fingerprint. This representation captures reaction-specific features, such as changes in molecular structures and functional groups.
    \item EC number representation: we encode them in a way that preserves their hierarchical relationships. Instead of treating whole EC numbers as simple categorical labels, which would ignore relationships between enzymes within the same category, we encode them as four separate numerical features, one for each EC digit. Each of these four digits is first label-encoded, then converted into a continuous representation via standardization, approaching it as a regression task where similar EC numbers remain closer in feature space.
\end{itemize}

For all tasks, EC number label encoding is done on the full set of EC numbers, while standardization is performed using only the training set statistics, preventing information leakage from the test set.

\subsubsection{Training and evaluation}

XGBoost models are trained separately for each task using the same training and test splits as the LLM experiments. We run the model for 100 boosting rounds and include early stopping to avoid overfitting. For the EC number prediction task, the problem is framed as a regression task with a squared loss, whereas for the other two tasks we use a logistic regression for the output bit-vector.
\begin{itemize}
    \item EC prediction task: the 1024-bit reaction fingerprint and the standardized, 4D vector of the encoded EC number, represent input and output respectively. Evaluation is done by reverting the standardization process for the prediction and checking whether the categorical encoding of the predicted EC digits matches the true labels exactly.
    \item Product and substrate prediction: the input is represented by a concatenation of the 256-bit Morgan fingerprint with the 4D encoding of the EC number, and the output is a 256-bit Morgan fingerprint. Since the fingerprints are binary, the output is considered correct if the generated fingerprint exactly matches the ground truth fingerprint, as an upper bound proxy of our "molecule matching" prediction task.
\end{itemize}
Since the EC number contributes with only four features to an input vector of hundreds of dimensions, we conducted additional experiments to explore its impact. Specifically we inflated the relative importance of the EC number by multiplying its four components by factors ranging from 5 to 100. We also completely removed the EC number from the input to test its effect on performance. Our tests show that the best performance is achieved by including the EC number with the default scaling factor of 1, confirming that enzymatic information contributes meaningfully to reaction prediction, even when it constitutes a small fraction of the feature space.

\end{document}